\documentclass[journal]{IEEEtran}
\usepackage{booktabs}
\usepackage{graphicx}
\usepackage{cite}
\usepackage{amsmath}
\usepackage{amssymb}
\usepackage{multirow}
\usepackage{color}
\ifCLASSINFOpdf
\else
\fi
\begin{document}
%\tableofcontents
\title{Video-based Person Re-identification with Accumulative Motion Context}
%\title{End-to-end Two-stream Recurrent Convolutional Network for Video-Based Person Re-identification}

\author{Hao~Liu, Zequn Jie, Karlekar Jayashree, Meibin Qi, Jianguo Jiang
	and Shuicheng Yan, \IEEEmembership{Fellow,~IEEE}, Jiashi Feng% <-this % stops a space
	\thanks{Hao Liu, Meibin Qi and Jianguo Jiang are with School of Computer and Information, Hefei University of Technology, P. R. China. Hao Liu is also with Department of Electrical and Computer Engineering, National University of Singapore, Singapore, e-mail: hfut.haoliu@gmail.com; qimeibin@163.com; jgjiang@hfut.edu.cn.}% <-this % stops a space
	\thanks{Karlekar Jayashree is with Panasonic R\&D Center Singapore, Singapore, e-mail: Karlekar.Jayashree@sg.panasonic.com. }
	\thanks{Zequn Jie, Shuicheng Yan and Jiashi Feng are with Department of Electrical and Computer Engineering, National University of Singapore, Singapore, e-mail: zequn.nus@gmail.com, eleyans@nus.edu.sg, elefjia@nus.edu.sg.}}%

\maketitle

\IEEEpubid{\begin{minipage}{\textwidth}\ \\[12pt] \centering
		 Copyright \copyright  20xx IEEE. Personal use of this material is permitted. However, permission to use this material for any other purposes must be obtained from the IEEE by sending an email to pubs-permissions@ieee.org.		
\end{minipage}}

\begin{abstract}
Video based person re-identification plays a central role in realistic security and video surveillance. 
In this paper we propose a novel Accumulative Motion Context  (AMOC)  network for addressing this important problem, which effectively exploits  the long-range motion context for robustly identifying the same person under challenging conditions. Given a video sequence of the same or different persons, the proposed AMOC network jointly learns appearance representation and motion context from a collection of adjacent frames using a two-stream convolutional architecture. Then AMOC accumulates clues from motion context by recurrent aggregation,   allowing effective information flow among adjacent frames and capturing dynamic gist of the persons.  The architecture of AMOC is end-to-end trainable and thus motion context can be adapted to complement appearance clues under unfavorable conditions (\textit{e.g.}, occlusions). Extensive experiments are conduced on three public benchmark datasets,  \textit{i.e.}, the iLIDS-VID, PRID-2011 and MARS datasets, to investigate the performance of AMOC. The experimental results demonstrate that the proposed AMOC network outperforms  state-of-the-arts for video-based re-identification significantly and confirm the advantage of exploiting long-range motion context for video based person re-identification, validating our motivation evidently.
\end{abstract}


% \fi
\IEEEpeerreviewmaketitle

\section{Introduction}
\IEEEPARstart{T}{he} person re-identification (re-id) problem has received increasing attention \cite{farenzena2010person, li2014deepreid, yi2014deep, ahmed2015improved, Chen2016An, Wang2015Cross, liu2016end, liu2015kernelized, yan2016person,mclaughlinrecurrent, wang2016person, you2016top, liu2015spatio, zhu2016video, gao2016temporally}, which associates different tracks of a person moving between non-overlapping cameras distributed at different physical locations. It aims to re-identify the same person captured by one camera in another camera at a new location. This task is essential to many important surveillance applications such as people tracking and forensic search. However, it is still a  challenging problem because pedestrian images captured in different camera views could display large variations in lightings, poses, viewpoints, and cluttered backgrounds. 

\IEEEpubidadjcol

In this work, we investigate the problem of video-based person re-identification. The recent  state-of-the-art methods often solve the person re-id task  by matching  spatial appearance features (\textit{e.g.} color and texture) using a pair of still images captured for persons~\cite{farenzena2010person, li2014deepreid, yi2014deep, ahmed2015improved, liu2016end, liu2015kernelized}.  However, appearance feature representations learned from still images of people are intrinsically limited due to the inherent visual ambiguity caused by, \textit{e.g.}, clothing similarity among people in public places and appearance variance from cross-camera viewing condition variations. In this work, we propose to explore both appearance and  motion information from video frame sequences of people for re-id.  The video setting is indeed more natural for re-id considering  a person is often captured by surveillance cameras in a video rather than only  a single image. By using sequences of persons' video frames, not only the spatial appearance but also the temporal information, such as persons' gait,  can be utilized to discern difficult cases when trying to recognize a person in another different camera. In general, describing person video can naturally be attributed to spatial and temporal cues. The spatial part  carries information about scenes and appearance information of the persons, like clothing color, height and shape, while the temporal part, in the form of motion across frames, conveys the movement of the observer (the cameras) and the person objects, which is complementary to the spatial part. 

Recently, spatial-temporal information has been explored in video-based person re-id works\cite{yan2016person,mclaughlinrecurrent, wang2016person, you2016top, liu2015spatio, zhu2016video, gao2016temporally}. In  \cite{mclaughlinrecurrent}, the authors extract optical flows between consecutive frames of a person to represent the short-term temporal information, then concatenate them with RGB images to construct the spatial-temporal features. However, this method simply uses one CNN to extract spatial and temporal information simultaneously. The limitation of current single-stream CNN architecture is that it cannot take  advantage of valuable temporal information and the performance is consequently limited by only using spatial (appearance) features. Thus the learned spatial-temporal features may not be sufficiently discriminative for identifying different persons in real videos.

\IEEEpubidadjcol

To overcome the limitation of single-stream CNN architecture, in related fields, such as action recognition (\cite{simonyan2014two, feichtenhofer2016convolutional}), the authors decompose spatial and temporal information processing into two individual streams to separately learn more representative features  from video sequences, and then the information from the two streams is fused at certain intermediate layers. For the person re-id task, in the process of discriminating two walking persons, at a specific location such as legs or arms of a person, they may have discriminative motion characteristics except for some spatial features such as trousers' colour or texture. As it can only be obtained between two consecutive frames, this motion is called motion context. Therefore, we design a novel two-stream spatial-temporal architecture, called the Accumulative Motion Context networks (AMOC), that can well capture the spatial features and motion context from sequences of persons. And then the useful appearance and motion information is accumulated  in a recurrent way for providing long-term information.

Furthermore, to our best knowledge, the motion information in most of related methods~\cite{simonyan2014two, feichtenhofer2016convolutional, mclaughlinrecurrent} is represented by hand-crafted optical flow extracted in an off-line way. Such off-line optical flow extraction has the following disadvantages: (i) The pre-extraction of optical flow is time and resources consuming due to the storage demanding, especially for the large-scale dataset, such as MARS~\cite{zheng2016mars}; (ii) Off-line extracted optical flow may not be optimal for the person re-id task since it is independent of person re-id. Instead, we expect the model to automatically learn motion information from video sequential frames discriminative for person re-id. To encode the motion information from raw video frames, a recent work called FlowNet\cite{dosovitskiy2015flownet} is proposed to estimate high quality optical flow. Basically, the idea of FlowNet is to exploit the ability of convolutional networks to learn strong features at multiple levels of scale and abstraction, which can help finding the actual correspondences between frames. The FlowNet consists of contracting and expanding parts trained as a whole using back-propagation. The optical flow is first learned and spatially compressed in contracting part and then refined to high resolution in expanding part. Inspired by FlowNet, we design a new motion network in our AMOC architecture to perform the end-to-end motion context information learning task. Although our proposed end-to-end AMOC also involves estimating the optical flow (encoding motion information) from raw video frames similar to FlowNet, there are still several distinct differences with that model, which are listed as follows: 
\begin{itemize} 
\item First, They have different targets. FlowNet is for computing low-level optical flow. In contrast, our proposed AMOC model is to utilize the flow information for performing high-level person re-identification tasks. Due to different application targets, the optimization goal of FlowNet is to minimize the endpoint errors (EPE)\cite{dosovitskiy2015flownet} for getting more precise optical flow while our AMOC is to jointly re-identify persons and estimate optical flow towards benefiting person re-id.   

\item Second, the motion net in our model works together with the spatial stream to learn both appearance feature and motion feature which are complementary to each other. By contrast, FlowNet only focuses on learning the motion information encoded by the optical flow.

\item  Last but not least, considering the data in person re-id datasets that we experimented on has much lower resolution than the synthetic data that FlowNet uses, the structure of relevant layers in motion net is essentially modified to allow our model to take in low resolution video frames, especially for the person video frames, as input.

\end{itemize}

Our experiments and analyses show our end-to-end trainable two-stream approach successfully learns motion features helpful for improving the performance of person re-identification. Moreover, we explore how to fuse the learned appearance feature and motion feature together can make our model perform best. Compared to other state-of-the-art video person re-id methods, our method is also able to achieve  better performance, as clearly validated by extensive experimental results. 

To sum up, we make the following contributions to video-based person re-identification:

\begin{itemize} 

\item We propose a novel two-stream motion context accumulating network model which can directly learn both spatial features and motion context from raw person video frames in an end-to-end manner, instead of pre-extracting optical flow using the off-line algorithm. Then the learned spatial appearance features and temporal motion context information are accumulated in a recurrent way.

\item We quantitatively validate the good performance of our end-to-end two-stream recurrent convolutional network by comparing it with the state-of-the-arts on three benchmark datasets: iLIDS-VID~\cite{wang2014person}, PRID-2011~\cite{hirzer2011person} and MARS~\cite{zheng2016mars}.
\end{itemize}

\section {Related work}
Person re-identification has been extensively studied in recent years. Existing works on person re-identification can be roughly divided into two types: person re-id for still images and  person re-id for video sequences.
\subsection{Person Re-id for Still Images}
Previous works on person re-id for still images focus on the invariant feature representation and distance metric learning. Discriminative features that are invariant to environmental and view-point changes play a determining  role in person re-id performance.~\cite{gray2008viewpoint} combines spatial and color information using an ensemble of discriminant localized features and classifiers selected by boosting to improve viewpoint invariance. Symmetry and asymmetry perceptual attributes are exploited in~\cite{farenzena2010person}, based on the idea that features closer to the body axes of symmetry are more robust against scene clutter.~\cite{cheng2011custom} fits a body configuration composed of chest, head, thighs, and legs in pedestrian images and extracts per-part color information as well as color displacement within the whole body to handle pose variation.~\cite{ma2012local} turns local descriptors into the Fisher Vector to produce a global representation of an image. Kviatkovsky \textit{et al.}~\cite{kviatkovsky2013color} propose a novel illumination-invariant feature representation based on the logchromaticity (log) color space and demonstrate that color as a single cue shows relatively good performance in identifying persons under greatly varying imaging conditions.

After feature extraction, distance metric learning is used in person re-id to emphasize inter-person distance and de-emphasize intra-person distance. Large margin nearest neighbor metric (LMNN)~\cite{weinberger2009distance} is proposed to improve the performance of the traditional kNN classification. Prosser \textit{et al.}~\cite{prosser2010person} formulate person re-identification as a ranking problem, and use ensembled RankSVMs to learn pairwise similarity. Zheng \textit{et al.}~\cite{zheng2013reidentification} propose a soft discriminative scheme termed relative distance comparison (RDC) by large and small distances corresponding to wrong matches and right matches, respectively.~\cite{liao2015person} proposes a high dimensional representation of color and Scale Invariant Local Ternary Pattern (SILTP) histograms. It constructs a histogram of pixel features, and then takes its maximum values within horizontal strips to tackle viewpoint variations while maintaining local discrimination.

\subsection{Person Re-id for Video Sequences}

Recently, some works consider performing person re-id in video sequences. In this scenario, the person matching problem in videos is crucial in exploiting multiple frames in videos to boost the performance.~\cite{simonnet2012re} applies Dynamic Time Warping (DTW) to solve the sequence matching problem in video-based person re-id.~\cite{karaman2012identity} uses a conditional random field (CRF) to ensure similar frames in a video sequence to receive similar labels.~\cite{li2013locally} proposes to learn to map between the appearances in sequences by taking into account the differences between specific camera pairs.~\cite{wang2014person} introduces a pictorial video segmentation approach and deploys a fragment selecting and ranking model for person matching.~\cite{karanam2015sparse} introduces a block sparse model to handle the video-based person re-id problem by the recovery problem in the embedding space.~\cite{liu2015spatio} proposes a spatio-temporal appearance representation method, and feature vectors that encode the spatially and temporally aligned appearance of the person in a walking cycle are extracted.~\cite{you2016top} proposes a top-push distance learning (TDL) model incorporating a top-push constraint to quantify ambiguous video representation for video-based person re-id.

Deep learning methods have also been applied to video-based person re-id to simultaneously solve feature representation and metric learning problem. Usually DNNs are used to learn ranking functions based on pairs~\cite{yi2014deep} or triplets~\cite{ding2015deep} of images. Such methods typically rely on a deep network (\emph{e.g.} Siamese network~\cite{hadsell2006dimensionality}) used for feature mapping from raw images to a feature space where images from the same person are close while images from different persons are widely separated.~\cite{mclaughlinrecurrent} uses a recurrent neural network to learn the interaction between multiple frames in a video and a Siamese network to learn the discriminative video-level features for person re-id.~\cite{yan2016person} uses the Long-Short Term Memory (LSTM) network to aggregate frame-wise person features in a recurrent manner. Unlike the existing deep learning based methods for person re-id in videos, our proposed Accumulative Motion Context (AMOC) networks introduces an end-to-end two-stream architecture that has specialized network streams  for learning spatial appearance and temporal feature representations individually. Spatial appearance information from the raw video frame input and temporal motion information from the optical flow predicted by the motion network are processed respectively and then fused at  higher recurrent layers to form a discriminative video-level representation.

\begin{figure*}[htb]
	\centering
	\includegraphics[height=9cm]{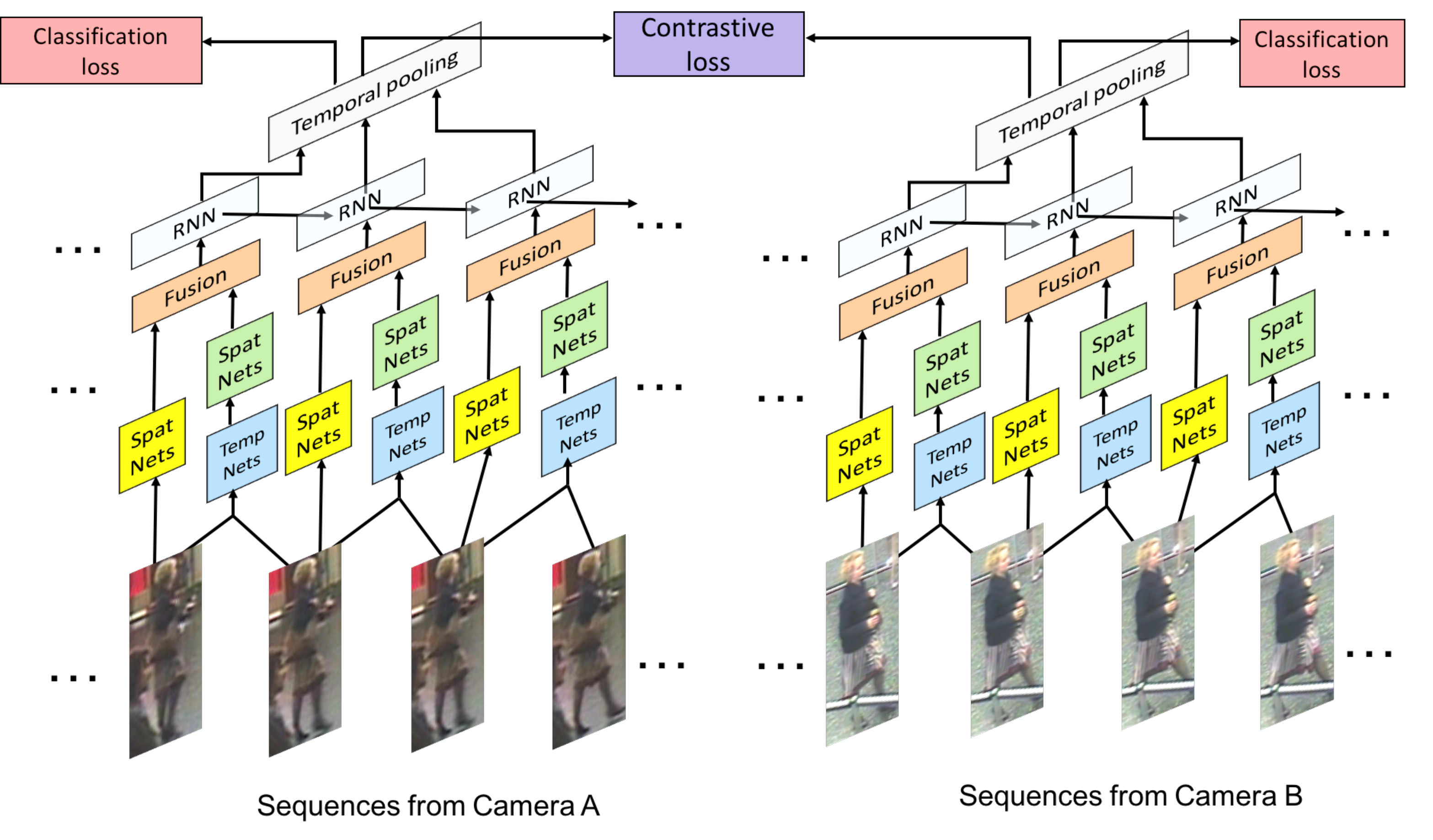}
	\vspace{-4.5mm}	
	\caption{The architecture of our proposed  Accumulative Motion Context Network (AMOC). At each time-step, each pair of two consecutive frames is processed by a two-stream network, including spatial network (yellow and green boxes) and motion network (blue box), to learn spatial appearance and temporal motion feature representations. Then these two-stream features are fused in a recurrent way for learning discriminative accumulative motion contexts. The two-steam features are integrated by temporal pooling layer from an arbitrarily long video sequence into a single feature representation. Finally, the two-stream sub-networks for two sequences from two different cameras are constructed following the Siamese network architecture\cite{hadsell2006dimensionality} in which the parameters of Camera A networks and Camera B networks are shared. The whole  AMOC network is end-to-end trained by introducing multi-task losses (classification loss and contrastive loss) to satisfy both the contrastive objective and to predict the person’s identity.}
	
	%\vspace{-5.5mm}	
	\label{fig:framework}
\end{figure*}
\section{Proposed Method}

We propose an end-to-end Accumulative Motion Context Network (AMOC) based architecture that addresses the video person re-identification problem through joint spatial appearance learning and motion context accumulating from raw video frames. We first introduce the overall architecture of our AMOC model (\ref{arch}), which is illustrated in Fig. \ref{fig:framework}. Then for each pair of frames from a person sequence, the details of motion networks and spatial networks  and how they collaborate with each other are described.  Besides, the recurrent fusion layers to integrate two-stream spatial-temporal information and fusion method are elaborated. Finally, implementation details of training and test are introduced for reproducing the results.
% An example of a floating figure using the graphicx package.
% Note that \label must occur AFTER (or within) \caption.
% For figures, \caption should occur after the \includegraphics.
% Note that IEEEtran v1.7 and later has special internal code that
% is designed to preserve the operation of \label within \caption
% even when the captionsoff option is in effect. However, because
% of issues like this, it may be the safest practice to put all your
% \label just after \caption rather than within \caption{}.
%
% Reminder: the "draftcls" or "draftclsnofoot", not "draft", class
% option should be used if it is desired that the figures are to be
% displayed while in draft mode.
%
%\begin{figure}[!t]
%\centering
%\includegraphics[width=2.5in]{myfigure}
% where an .eps filename suffix will be assumed under latex, 
% and a .pdf suffix will be assumed for pdflatex; or what has been declared
% via \DeclareGraphicsExtensions.
%\caption{Simulation results for the network.}
%\label{fig_sim}
%\end{figure}

\subsection{Architecture Overview }\label{arch}
Fig.~\ref{fig:framework} illustrates the architecture of the proposed end-to-end Accumulative Motion Context network (AMOC). In our architecture each two consecutive frames is processed by a two-stream network to learn spatial appearance and temporal motion features representing the person's appearance at a certain time instant. Then these two-stream features are fused in a recurrent way for learning discriminative accumulative motion contexts. After that, the temporal pooling layer integrates the two-steam features in time from an arbitrarily long video sequence into a single feature representation. Finally, the two-stream sub-networks for two sequences from two different cameras are constructed following the Siamese network architecture~\cite{hadsell2006dimensionality} in which the parameters of Camera A networks and Camera B networks are shared. To end-to-end train this network, we adopt multi-task loss functions including  
contrastive loss and classification loss. The contrastive loss decides whether two sequences describe the same person or not while the classification loss predicts the identity of the person in the sequence. In the following, we will give more detailed explanations to each component of our proposed network. 

% Note that the IEEE typically puts floats only at the top, even when this
% results in a large percentage of a column being occupied by floats.
\subsection{End-to-end Two-stream  Networks }
As aforementioned,  each frame in a sequence of the person is processed by  two convolutional network streams jointly. Specifically, one stream uses spatial networks (yellow boxes in Fig. \ref{fig:framework}) to learn spatial features from raw video frames while for the other stream spatial networks (green boxes in Fig.~\ref{fig:framework}) are applied on the motion context features produced by motion networks (blue boxes in Fig.~\ref{fig:framework}) to learn the temporal feature representations at each location between a pair of video frames. In this subsection, we introduce the details of the motion networks and spatial networks, and then describe how they work together. 
\subsubsection{The Motion Networks}\label{motion_net}
\begin{figure*}[htb]
	\centering
	\includegraphics[width=16cm]{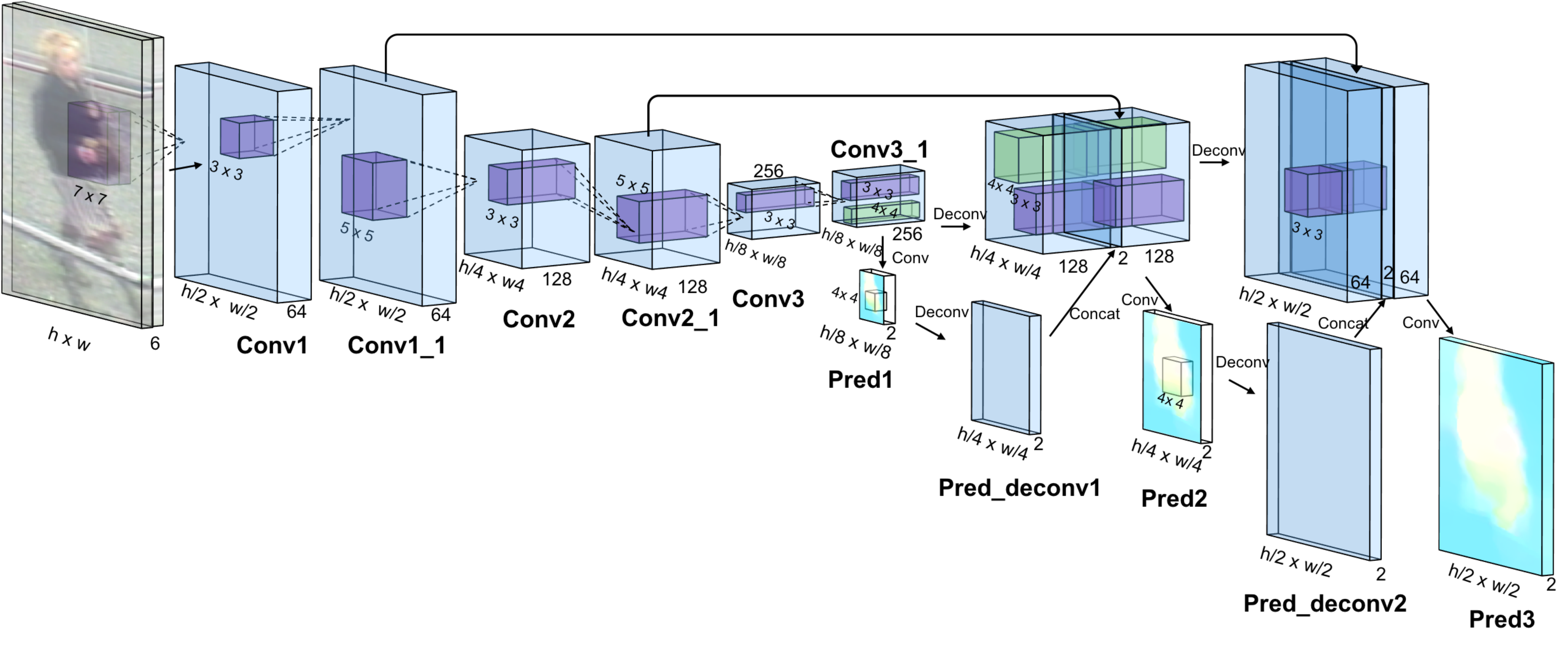}
	\vspace{-3.5mm}	
	\caption{The structure of motion networks of our proposed accumulative motion context network at one time-step.  It has 6 convolutional layers (corresponding to ``Conv1'', ``Conv\_1'', ``Conv2'', ``Conv2\_1'',``Conv3'' and ``Conv3\_1'' ) with stride of 2 in six of them and a tanh non-linearity layer after each layer. The inputs are the concatenated two person frames with size of $h \times w \times 6$. To provide dense per-pixel predictions, several deconvolutional layers are applied on output feature maps of convolutional layers and motion predictions to refine the coarse pooled representation. The purple cubes represent the convolutional kernels while the green cubes are the deconvolutional kernels. }
	
	%\vspace{-5.5mm}	
	\label{fig:flownet}
\end{figure*}

As shown in Fig.~\ref{fig:flownet}, at each time-step a pair of consecutive video frames of a person is processed by the motion network within AMOC (corresponding to blue boxes in Fig. \ref{fig:framework}) to predict motion between the adjacent frames. Similar to the structure used in \cite{dosovitskiy2015flownet}, the motion network consists of several convolutional and  deconvolutional layers for up-sizing the high-level coarse feature maps learned by convolutional layers. In particular, it has 6 convolutional layers (corresponding to ``Conv1'', ``Conv\_1'', ``Conv2'', ``Conv2\_1'',``Conv3'' and ``Conv3\_1'' ) with stride of 2 (the simplest form of pooling) in six of them and a tanh non-linearity after each layer.  Taking the concatenated two person frames as inputs with size of $h \times w \times 6$ (\textit{h} is the height of the frame and \textit{w} is the width), the network employs 6 convolutional layers  to learn high-level abstract representations of the frame pair by producing the feature maps with reduced sizes (\textit{w.r.t.} the raw input frames). However, this size shrinking process could result in low resolution and harm the performance of person re-id. So in order to provide dense per-pixel predictions we need to refine the coarse pooled representation.  To perform the refinement, we apply the ``deconvolution" on feature maps, and concatenate them with corresponding feature maps from the ``contractive" part of the network and an upsampled coarser flow prediction (``Pred1'', ``Pred2'', ``Pred3''). In this way, the networks could preserve both the high-level information passed from coarser feature maps and refine local information provided in lower layer feature maps. Each step increases the resolution by a factor of 2. We repeat this process for twice, leading to the final predicted flow (``Pred3'') whose resolution is still as half as that of raw input. Note that our motion networks do not have any fully connected layers, which can take video frames of arbitrary size as input. This motion network can be end-to-end trained by using optical flow generated by several off-the-shelf algorithms as supervision, such as Lucas-Kanade \cite{lucas1981iterative} and EpicFlow\cite{revaud2015epicflow} algorithm. The training details will be described in Sec. \ref{pre_motion}.

\subsubsection{The Spatial Networks}\label{spatial}
\begin{figure}[htb]
	\centering
	\includegraphics[width=8.5cm]{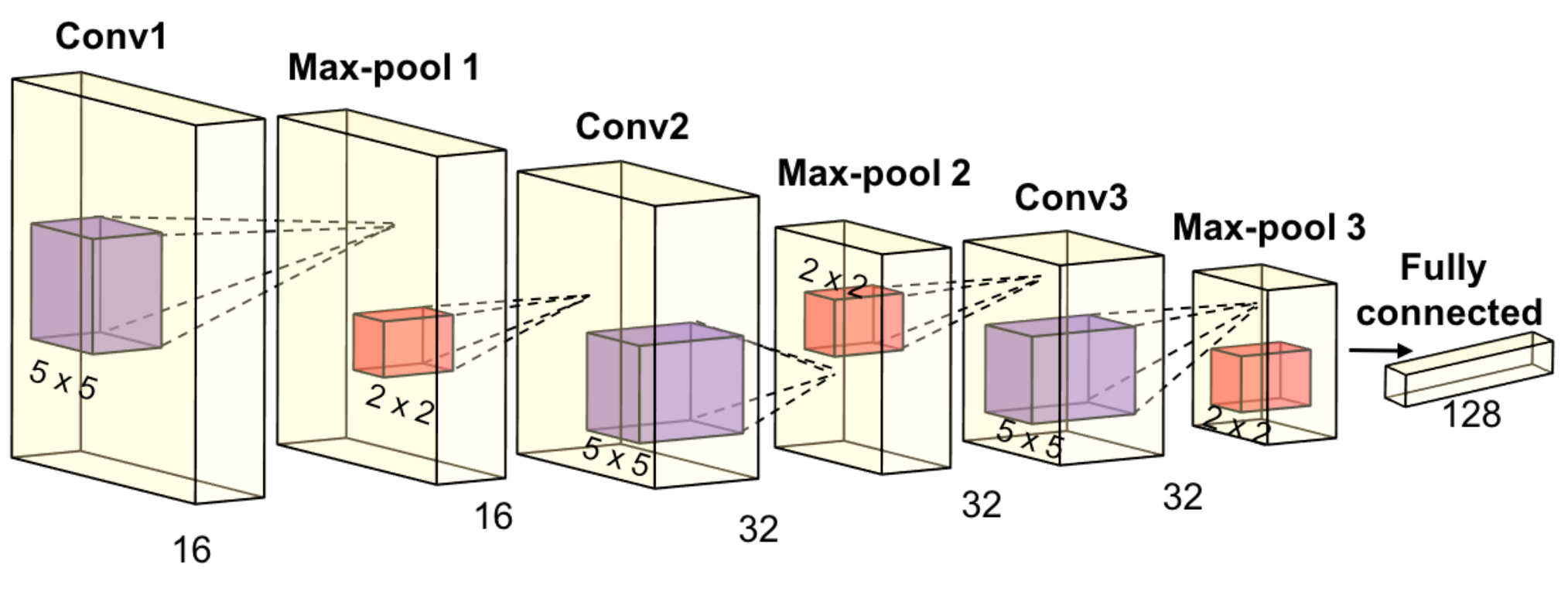}
	\vspace{-3.5mm}	
	\caption{The structure of spatial networks of our proposed accumulative motion context network at one time-step. It consists of 3 convolutional layers and 3 max-pooling layers with a \textit{tanh} non-linearity layer interpolated after each convolutional layer. And there is a fully-connected layer at the top of last max-pooling layer. The purple cubes are convolutional kernels while the red ones are pooling kernels.}
	
	%\vspace{-5.5mm}	
	\label{fig:spatnet}
\end{figure}
As shown in Fig. \ref{fig:framework}, there are two spatial networks (yellow and green boxes) in both streams to learn spatial feature representations from raw video frames and temporal features at the spatial location between two consecutive frames. Here, both spatial networks lying in the two streams have the same structure, each of which contains  3 convolutional layers and 3 max-pooling layers with a non-linearity layer (in this paper we use \textit{tanh}) interpolated after each convolutional layer. And there is a fully-connected layer at the top of last max-pooling layer. The details of spatial networks are shown in Fig. \ref{fig:spatnet}, where the purple cubes are convolutional kernels and the red ones are pooling kernels. All the strides in the convolutional layers and pooling layers are set as 2. Note that although two spatial networks have the same structure, they play different roles in two streams. The inputs of the network (yellow boxes in Fig.~\ref{fig:framework}) are raw RGB images when serves as a spatial feature extractor. Otherwise, the inputs are the last predictions (``Pred3'') of motion networks (blue boxes in Fig. \ref{fig:framework}). 

%\subsubsection{Spatial-Temporal Fusion}

\subsection{Spatial Fusion and Motion Context Accumulation}

\subsubsection{Spatial fusion}\label{sp_fuse}

Here we consider different fusion methods (orange boxes in Fig. \ref{fig:framework} ) to fuse the two stream networks. Our intention is to fuse the spatial features and motion context information at the spatial location  such that channel responses at the same pixel position are put in correspondence. Because the structures of spatial networks in two streams are the same, the feature map produced by each layer in each stream has the exact location correlation. So the problem is how to fuse the outputs of corresponding layers of two streams. To motivate our fusion strategy, consider for example discriminating two walking persons. If legs move or arms swing periodically at some spatial location then the motion network can recognize that motion and obtain the motion context information from two consecutive frames, and the spatial network can recognize the location (legs or hands) and their combination so as to discriminate the persons.

This spatial correspondence can be easily obtained when the two networks have the same spatial resolution at the layers to be fused. The  simplest way is overlaying (stacking) layers from one network on the other. However, there is also the issue of establishing the correct correspondence between one channel (or channels) in one network and the corresponding channel (or channels) of the other network. Suppose different channels in the spatial network learning spatial feature representation from one video frame are responsible for different body areas (head, arms, legs, \textit{etc.}), and one channel in the spatial network following the motion network is responsible for contextual motion between two neighboring frames in the fields. Then, after the channels are stacked, the filters in the subsequent layers must learn the correspondence between these appropriate channels  in order to best discriminate between these motions from different person samples.
To make this more concrete, we investigate the following 3 ways of fusing layers between two stream networks. Suppose  $\textbf{x}^{A} \in \mathbb{R}^{H \times W \times D}$ and $\textbf{x}^{B} \in \mathbb{R}^{H \times W \times D}$ are two feature maps from two layers and need to be fused, where \textit{W}, \textit{H} and \textit{D} are the width, height and channel number of the respective feature maps. And $\textbf{y} $ represents the fused feature maps. When applied to the feedforward spatial network architecture which is shown in Fig. \ref{fig:spatnet}, which consists of convolutional, max-pooling non-linearity and fully-connected layers, the fusion can be performed at different points in the network to implement \textit{e.g.} early-fusion or late-fusion. 
\begin{itemize}
	\item \textbf{Concatenation fusion}  This fusion operation stacks the two feature maps at the same spatial locations \textit{i}, \textit{j} across the feature channels \textit{d}:

\begin{equation}\label{concf}
y_{i,j,2d}^{\mathrm{cat}}= x_{i,j,d}^{\mathrm{A}} ,  y_{i,j,2d-1}^{\mathrm{cat}} =  x_{i,j,d}^{\mathrm{B}},
\end{equation}		
where $\textbf{x}^{A}, \textbf{x}^{B}\in\mathbb{R}^{H \times W \times D} , \textbf{y}^{\mathrm{cat}} \in \mathbb{R}^{H \times W \times 2D}$ and $1\leq i \leq H,1\leq j \leq W$.
	\\
	\item \textbf{Sum fusion} The sum fusion computes the sum of the two feature maps at the same spatial locations \textit{i}, \textit{j }and channels \textit{d}:
\begin{equation}\label{sumf}
y_{i,j,d}^{\mathrm{sum}}= x_{i,j,d}^{\mathrm{A}} + x_{i,j,d}^{\mathrm{B}},
\end{equation}		
	where $ \textbf{y}^{\mathrm{cat}} \in \mathbb{R}^{H \times W \times 2D}$. 
	\item \textbf{Max fusion}
	Similarly, max fusion takes the maximum of the two feature maps:
	\begin{equation}\label{convf}
	y_{i,j,d}^{\mathrm{max}}=\mathrm{max} \{ x_{i,j,d}^{\mathrm{A}} , x_{i,j,d}^{\mathrm{B}} \} .
	\end{equation}
\end{itemize}	

\begin{figure}[htb]
	\centering
	\includegraphics[width=7.5cm]{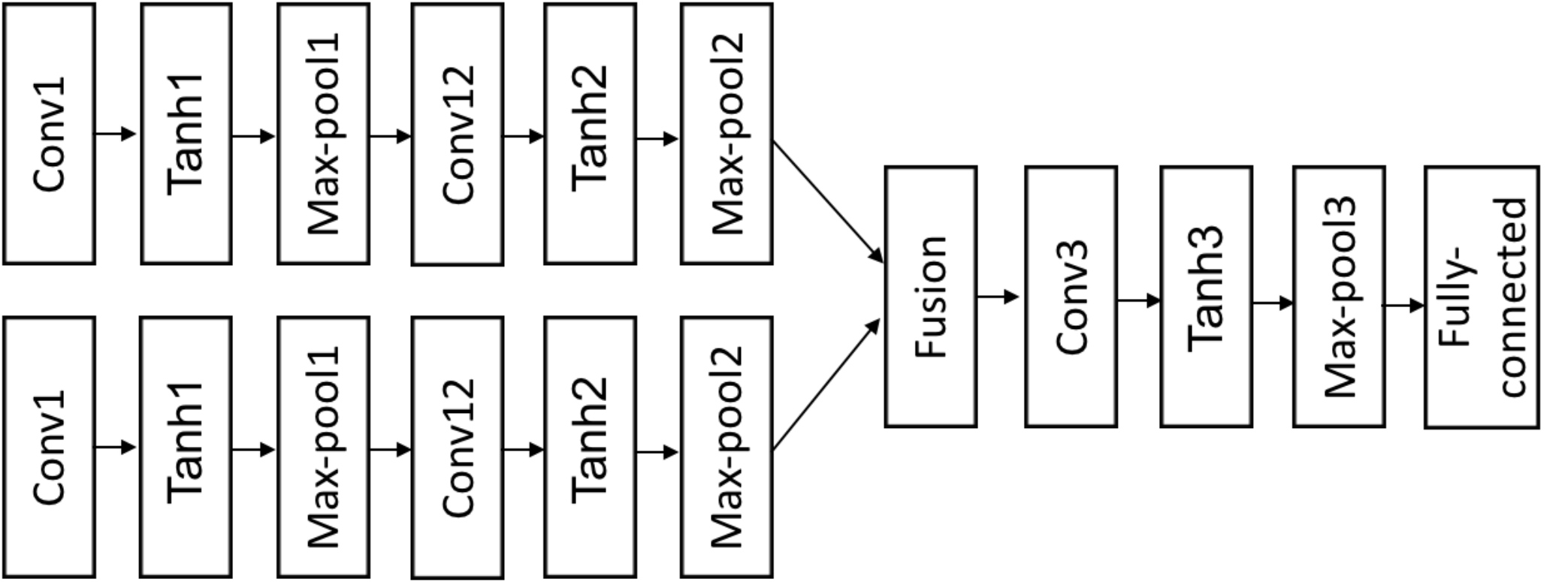}
	%\vspace{-5.5mm}	
	\caption{The illustration of fusion of two spatial networks at the second max-pooling layer (Max-pool2).}
	
	%\vspace{-5.5mm}	
	\label{fig:fusion}
\end{figure}

Now, we briefly introduce where our fusion method should be applied to. Injecting fusion layers can have significant impact on the number of parameters and layers in a two-stream network, especially if only the network which is fused into is kept and the other network tower is truncated. For example, in Fig. \ref{fig:spatnet}, if the fusion operation is performed on the second max-pooling layers  (``Max-pool2'') of two spatial networks in the two-stream network, then the previous parts (``Conv1, Max-pool1, Conv2'' and non-linearity layers) are kept, and the layers (``Conv3, Max-pool3, Fully connected'') after fusion operation share one set of parameters. The illustration is shown in Fig. \ref{fig:fusion}.
	
In the experimental section (Sec. \ref{ana}), we evaluate and compare the performance of each of the three fusion methods in terms of their re-identification accuracy.
\subsubsection{Motion Context Accumulation}\label{mo_acc}
We now consider the techniques to combine fused features  $\textbf{f}^{(t)}$ (output of the fully-connected layer in Fig. \ref{fig:fusion}) containing both spatial appearance and motion context information over time \textit{t}. Because the length of a sequence is arbitrary, the motion context is also unfixed for each person. Therefore, we exploit  Recurrent neural networks (RNN) which can process an arbitrarily long time-series using a neural network to address the problem of accumulating motion context information. Specifically, an RNN has feedback connections allowing it to remember information over time and produces an output based on both the current input and information from the previous time-steps at each time-step.  As shown in Fig. \ref{fig:framework},  the recurrent connections of RNN are ``unrolled"  in time to create a very deep feed-forward network.  Given the unrolled network, the lateral connections serve as ``memory'', allowing information to flow between an arbitrary number of time-steps. 

As video-based person re-identification involves recognizing a person from a sequence, accumulating the motion context information of each frame at each instant would be helpful to improve the performance of re-identification. This motion accumulation can be achieved by using the recurrent connections allowing information to be passed between time-steps. To be more clear,  we aim to better capture person spatial appearance features and motion context information present in the video sequence, and then to accumulate them along the time axis. Specifically, given the \textit{p}-dimension output $\textbf{f}^{(t)} \in \mathbb{R}^{p \times 1}$ of the fused spatial networks, the RNN can be defined as follows:
\begin{align}\label{RNN}
\textbf{o}^{(t)}&= M\textbf{f}^{(t)} + N\textbf{r}^{(t-1)},\\
\textbf{r}^{(t)}&= tanh(\textbf{o}^{(t)}).
\end{align}
Here $\textbf{o}^{(t)} \in \mathbb{R}^{q \times 1}  $ is the $q$-dimensional output of RNN at time-step $t$, and $\textbf{r}^{(t-1)} \in \mathbb{R}^{q \times 1}  $  contains the information on the RNN's state at the previous time-step. The $M\in \mathbb{R}^{q \times p}$ and $N\in \mathbb{R}^{q \times q}$ represent the corresponding parameters for $\textbf{f}^{(t)}$ and $\textbf{r}^{(t-1)}$ respectively, where $q$ is the dimension of the output of the last fully-connected layer in fusion part and $p$ is the dimension of the feature embedding-space. 

Although RNNs are able to accumulate the fused information, they still have some limitations. Specifically, some time-steps may be more dominant in the output of the RNN, which could reduce the RNN’s effectiveness when used to accumulate the input information over a full sequence  because discriminative frames may appear anywhere in the sequence. To overcome the drawback, similar to \cite{mclaughlinrecurrent}, we add a temporal pooling layer after RNN to allow for the aggregation of information across all time steps.  The temporal pooling layer aims to capture long-term information present in the whole sequence, which combines with the motion context information accumulated through RNN. In this paper, we can use either average-pooling  or max-pooling over the temporal dimension to produce a single feature vector. $\textbf{u}_{ave}$ represents the person's spatial appearance and motion information averaged over the whole input sequence while $\textbf{u}_{max}$ denotes the feature vector after max-pooling. Given the temporal pooling layer inputs $\left\lbrace \textbf{o}^{(1)}, \textbf{o}^{(2)}, ...,  \textbf{o}^{(T)} \right\rbrace$, the average-pooling and max-pooling methods are as follows:
\begin{align}
\textbf{u}_{ave} &=\dfrac{1}{T}\sum_{t=1}^{T}\textbf{o}^{(t)},\\
\textbf{u}_{max}^i &=\mathrm{max}\left( \left[ \textbf{o}^{(1),i},\textbf{o}^{(2),i},...,\textbf{o}^{(t),i}\right] \right), 
\end{align}
where $T$ is the length of the sequence or time-steps. And $\textbf{u}_{max}^i$  is the i’th element of the vector vs and $\left[ \textbf{o}^{(1),i},\textbf{o}^{(2),i},...,\textbf{o}^{(t),i}\right]$ are \textit{i}th elements of the feature vector across the temporal dimension.
\subsubsection {Multi-task Loss}
Similar to the method suggested in~\cite{mclaughlinrecurrent}, we train the whole  AMOC network to satisfy both the contrastive objective and to predict the person’s identity. Given the sequence feature vector \textbf{u} including accumulative spatial appearance feature and motion context information, output by the  feature representation learning networks of our AMOC, we can predict the identity of  the person in the sequence using the standard softmax function, which is defined as follows:
\begin{equation}
\textit{I }(\mathbf{u}) = P(z=c|\mathbf{u})=\frac{\exp(S_{c}\mathbf{u})}{\sum_{k}\exp(S_{k} \mathbf{u})},
\end{equation}
where there are a total of \textit{K} identities, \textit{z} is the identity of the person, and \textit{S} is the softmax weight matrix while $S_{c}$ and $S_{K}$ represent the \textit{c}th and \textit{k}th column of it, respectively. Then the corresponding softmax loss function (pink boxes in Fig. \ref{fig:framework}) is defined as follows:
\begin{equation}
L_{class} = -\textup{log}(P(z=c|\mathbf{u})).
\end{equation}

Besides, given a pair of sequence feature vectors $(\mathbf{u}_{(a)}, \mathbf{u}_{(b)})$  output by the Siamese network,  the contrastive loss (purple box in Fig.\ref{fig:framework}) function can be defined as
\begin{align}\label{con_loss}
L_{con} = \left \| \mathbf{u}_{(a)} - \mathbf{u}^+_{(b)}  \right \|_{2}^{2}+\mathrm{max}\left\{0, \alpha-  \left \| \mathbf{u}_{(a)} - \mathbf{u}^-_{(b)}  \right \|_{2}^{2} \right\},
\end{align}
where $\mathbf{u}^+_{(b)}$ represents the positive pair of $\mathbf{u}_{(a)}$ while $\mathbf{u}^-_{(b)}$ represents the negative pair of $\mathbf{u}_{(a)}$. The loss consists of two penalties: the first term penalizes a positive pair $(\mathbf{u}_{(a)}, \mathbf{u}^+_{(b)})$ that is too far apart, and the second penalizes a negative pair $(\mathbf{u}_{(a)}, \mathbf{u}^-_{(b)})$ that is closer than a margin $\alpha$. If a negative pair is already separated by $\alpha$, then there is no penalty for this pair and $L_{con}(\mathbf{u}_{(a)},\mathbf{u}^-_{(b)}) = 0$.

Finally, we jointly end-to-end train our architecture with both classification loss and contrastive loss.  We can now define the overall multi-task training loss function  $\textit{L}_{multi}$ for a single pair of person sequences, which jointly optimizes the classification cost and the contrastive cost as follows:
\begin{align}
L_{multi}(\mathbf{u}_{(a)},\mathbf{u}_{(b)}) = &L_{con}(\mathbf{u}_{(a)},\mathbf{u}_{(b)})\notag\\
&+L_{class}(\mathbf{u}_{(a)})+L_{class}(\mathbf{u}_{(b)}).
\end{align}
Here, we give equal weights for the classification cost and contrastive cost terms. The above network can be trained end-to-end using back-propagation-through-time from raw video frames (details of our training parameters can be found in Sec. \ref{impl_details}). During the training phase, all recurrent connections are unrolled to create a deep feed-forward graph, where the weights of AMOC are shared between all time-steps. And in the test phase,  we discard the multi-task loss functions and apply the AMOC on the raw video sequences as a feature extractor, where the feature vectors extracted by it can be directly compared using Euclidean distance.

\subsection{Implementation Details}\label{impl_details}
\subsubsection{Pre-training of the Motion Networks}\label{pre_motion}
As aforementioned in the Sec.~\ref{motion_net}, we use the pre-extracted optical flow as the ground truth to pre-train the motion networks. Our aim is that our proposed architecture can directly learn motion context information from raw consecutive video frames. When the pre-training of motion networks is finished, we use the trained parameters to initialize the motion networks in the whole framework. 

As shown in the Fig.~\ref{fig:flownet}, our motion networks produce three motion maps of optical flow maps at three levels of scale (``Pred1'', ``Pred2'', ``Pred3'') for a pair of person frames. Because the size of each prediction map is 1/8, 1/4 and 1/2 of the input frame size respectively, the pre-extracted optical flow maps are all downsampled to the corresponding sizes, then serve as the ground truth. In this paper, we use the smooth-$L_{1}$  loss function~\cite{girshick2015fast} to compute the losses between the predictions $\textbf{e}^{(l)} \in\mathbb{R}^{h^{(l)}\times w^{(l)} \times 2} $ and ground truth optical flow maps $\textbf{g}^{(l)} \in\mathbb{R}^{h^{(l)}\times w^{(l)} \times 2} $, where $l = 1, 2 ,3$. The loss function is defined as

\begin{align}
L^{(l)}_{(motion)}(\textbf{e}^{(l)}, \textbf{g}^{(l)})&=\sum_{i,j,k}\textup{smooth}_{L_{1}}(e_{i,j,k}^{(l)}- g_{i,j,k}^{(l)}),\\
\textup{smooth}_{L_{1}}(\theta)&=\left\{\begin{matrix}
&0.5\theta^{2}   &\textup{if}\left | \theta \right |<1\\ 
&\left |\theta \right |-0.5&\textup{otherwise}.
\end{matrix}\right.
\end{align} 
Then, the overall cost function can be written as
\begin{equation}
L_{(motion\_all)}=\sum_{l=1}^{3}\omega_{l}L^{(l)}_{(motion)},
\end{equation}
where $\omega_{l}$ represents the weight of each loss between different scale level prediction and ground truth map. In the pre-training phase of the motion network, we set them to (0.01, 0.02, 0.08) respectively. 

 All the input pairs of video frames are resized to $128 \times 64$ and Adam\cite{kingma2014adam} is chosen as the optimization method due to its faster convergence than standard stochastic gradient descent with momentum for our task. As recommended in \cite{kingma2014adam}, we fix the parameters of Adam : $ \beta_{1}= 0.9$ and $\beta_{2} = 0.999$. Considering every pixel is a training sample, we use small mini-batches of 4 frame pairs. We start with learning rate $\lambda= 1e-4$ and then divide it by 2 every 10k iterations after the first 20k. Note, the motion network of our model has a good generalization ability, which is verified in the following Section (Sec. \ref{ana}). Therefore, in the experiments on all three datasets, we use the motion network only pre-trained on the iLIDS-VID to initialize the one in the whole end-to-end AMOC framework.
 
\subsubsection{Training of the Overall Architecture}
After pre-training the motion network, we use it to initialize the one in our AMOC to end-to-end train the whole network. In the end-to-end training process, we set margin $\alpha$ to 2  in the Eqn. \eqref{con_loss}, and the embedding-space dimension is set to 128. For the temporal pooling, we adopt the average-pooling method in the related experiments if not specified. Besides, the learning rate is set to $1e-3$. Note, as we mentioned in  Sec. \ref{motion_net}, the resolution of output final predicted flow map of the motion network (``Pred3'')  is as half as that of input. Therefore, at each time-step, we resize the first frame of a pair to two scales of $64 \times 32$ and $128 \times 64$. For the second frame, we resize it to $128 \times 64$. That is to say, at each time-step, the first-stream spatial network (yellow boxes in Fig.\ref{fig:framework}) takes the first frame within a pair with size of $64 \times 32$ to learn spatial appearance feature representations. And the second-stream networks, including spatial network and motion network (green and blue boxes in Fig.\ref{fig:framework}) take the pair frames with sizes of $128 \times 64$ as input to extract motion context information. This operation can be performed in an on-line way. Furthermore, it guarantees that the feature map generated by each layer of two spatial networks in two streams has the same resolution thus can be fused at an arbitrary layer. 
\subsubsection{Data Augmentation}
To increase diversity of the training sequences to overcome the data imbalance and overfitting issue, data augmentation is applied on all the datasets. Specifically, we artificially augment the data by performing random 2D translation, similar to the processing in \cite{mclaughlinrecurrent}. For all frame images of size $w \times h$ in a sequence, we sample same-sized frame images around the image center, with translation drawn from a uniform distribution in the range $[\text{-0.05}w, \text{0.05}w] \times [\text{-0.05}h, \text{0.05}h]$. Besides, the horizontal flip is also performed to augment the data. At each epoch of training phase, the augmentation is applied once. During the testing phase, data augmentation is also applied, and the similarity scores between sequences are averaged over all the augmented data.
\section{Experiments}
\subsection{Datasets }
 In this paper, we use iLIDS-VID\cite{wang2014person}, PRID-2011\cite{hirzer2011person} and MARS\cite{zhu2016video}, which are three public benchmarks available, to evaluate our proposed AMOC model. 
 
 \noindent\textbf{iLIDS-VID}:$\quad$The iLIDS-VID dataset is created from the pedestrians captured in two non-overlapping camera views at an airport arrival hall under a multi-camera CCTV network. It is very challenging due to large clothing similarities among people, lighting and viewpoint variations across camera views, cluttered background and random occlusions. There are 600 image sequences of 300 distinct individuals in the dataset, with one pair of image sequences from two camera views for each person. The image sequences with an average number of 73 range in length from 23 to 192 image frames. 
 
 \noindent\textbf{PRID-2011}:$\quad$The PRID-2011 dataset  totally contains 749 persons captured by two non-overlapping cameras. Among,  there are 400 image sequences for 200 people from two camera views that are adjacent to each other, with sequences lengths of 5 to 675 frames. Compared with the iLIDS-VID dataset, it is less challenging due to being captured in non-crowded outdoor scenes with relatively simple and clean backgrounds and rare occlusions. Similar to the protocol used in \cite{mclaughlinrecurrent}, we only use the first 200 persons appearing in both cameras for evaluation.

 \noindent\textbf{MARS}:$\quad$ The MARS containing 1,261 IDs and around 20,000 video sequences is the largest video re-id  benchmark dataset to date.  Each sequence is automatically obtained by the Deformable Part Model (Deformable Part Model)\cite{Felzenszwalb2014Object} detector and the GMMCP\cite{Dehghan2015GMMCP} tracker. These sequences are captured by six cameras at most and two cameras at least, and each identity has 13.2 sequences on average. Additionally, there are also 3,248 distractor sequences are contained in the dataset. The dataset is fixedly split into training and test sets, with 631 and 630 identities, respectively. In testing, 2,009 probes are selected for query.
 \begin{table*}[htbp]
 	\linespread{1.5}\selectfont
 	\centering
 	\caption{Rank1, Rank5, Rank10 and Rank20 recognition rate (in \%) of various  methods on iLIDS-VID and PRID-2011 datasets. }
 	\begin{tabular}{c|cccc|cccc}
 		\hline
 		\textbf{Dataset} &\multicolumn{4}{c|}{\textbf{iLIDS-VID}}&\multicolumn{4}{c}{\textbf{PRID-2011}}\\
 		\hline
 		\textbf{Methods} & \textbf{Rank1} & \textbf{Rank5} & \textbf{Rank10} & \textbf{Rank20} & \textbf{Rank1} & \textbf{Rank5} & \textbf{Rank10} & \textbf{Rank20}\\
 		\hline
 		Baseline + LK-Flow\cite{mclaughlinrecurrent} & 58.0 & 84.0 & 91.0 & 96.0 & 70.0 & 90.0 & 95.0 & 97.0\\
 		Baseline + EpicFlow\cite{revaud2015epicflow} & 59.3 & 87.2 & 92.7 & 98.2 &76.2 & 97.5&98.2 &99.0\\ 
 		AMOC w/o Motion		                                    & 54.2 & 78.3 & 89.1 & 95.8 &65.4 & 88.9 &95.6 &98.5\\ 
 		AMOC + LK-Flow & 63.3 & 85.3 & 95.1 & 96.4 &76.0 & 96.5&97.4 &99.6 \\
 		AMOC + EpicFlow & 65.5& 93.1& 97.2 & 98.7 &82.0 &97.3 &99.3 &99.4\\
 		end-to-end AMOC + LK-Flow & 65.3 & 87.3 & 96.1 & 98.4 &78.0 &97.2 &99.1 &99.7 \\
 		\hline
 		\textbf{end-to-end AMOC + EpicFlow} & \textbf{68.7} & \textbf{94.3} & \textbf{98.3} & \textbf{99.3} & \textbf{83.7} & \textbf{98.3} & \textbf{99.4} & \textbf{100} \\
 		\hline
 	\end{tabular}%
 	
 	\label{tab:lk_epic}%
 \end{table*}%
 \subsection{Experimental Settings and Evaluation Protocol}\label{exp_setting}
 For the experiments performed on iLIDS-VID and PRID-2011 datasets,  half of persons are extracted for training and the other half for testing. All experiments are conducted 10 times with different training/testing splits and the averaged results ensure their stability. For the MARS dataset, we use the provided fixed training and test set, containing 631 and 630 identities respectively, to conduct experiments. Additionally, the training of networks in our architecture, including pre-training of motion networks and end-to-end training of the whole framework, are all implemented by using the Torch\cite{torch} framework on NVIDIA GeForce GTX TITAN X GPU.
 
 As we described in Sec. \ref{arch}, our network is a Siamese-like network, so the positive  and negative sequence pairs are randomly on-line selected during the training phase. However, positive and negative sequence pairs consist of two full sequences of an arbitrary length containing the same person or different persons under different cameras respectively. To guarantee the fairness of experiments, we follow the same sequence length setting in \cite{mclaughlinrecurrent}. Considering the efficiency of  training,  a sub-sequence containing 16 consecutive frames is sampled from the full length sequence of a person during training. We train our model for 1000 epochs on the iLIDS-VID and PRID-2011 datasets. On the MARS, the epoch number is set to 2000. At each epoch, the random selection is performed once. During testing, the sequences under the first camera and second camera are regarded as the probe and  the gallery respectively, as in \cite{mclaughlinrecurrent} and \cite{wang2014person}. And the length of each person sequence is set to 128 for testing. If the length of a full sequence of  a person is smaller than 128, we use the full-length sequence of this person to test. For the iLIDS-VID and PRID-2011 datasets, the training takes approximately 4-5 hours. And the training on the MARS consumes around 1 day. In the test phase, it takes roughly 5 minutes for one split testing on the iLIDS-VID and PRID-2011 datasets. On the MARS with larger scale, the testing time is increased to about 15 minutes.
 
 In experiments, for each pedestrian, the matching of his or her probe sequence (captured by one camera) with the gallery sequence (captured by another camera) is ranked. To reflect the statistics of the ranks of true matches, the Cumulative Match Characteristic (CMC) curve is adopted as the evaluation metric.  Specifically, in the testing phase, the Euclidean distances between probe sequence features and those of gallery sequences are computed firstly. Then, for each probe person, a rank order of all the persons in the gallery is sorted from the one with the smallest distance to the biggest distance. Finally, the percentage of true matches founded among the first $m$ ranked persons is computed and denoted as rank$m$. In addition, for the MARS dataset, the mean average precision (mAP) as in \cite{zheng2016mars} is also employed to evaluate the performance since each query has multiple cross-camera ground truth matches. Note, all the experiments performed on three datasets are under single-query setting.

\subsection{Analysis of the Proposed AMOC Model}\label{ana}
 
Before showing the comparison of our method with the state-of-the-arts, we conduct several analytic experiments on iLIDS-VID and PRID-2011 datasets to verify the effectiveness of our model for solving the video-based person re-identification problem. We analyse  and investigate the effect of several factors upon the performance, which include the  generation of motion context information, the selection of spatial fusion method, the location of performing spatial fusion, the sequence lengths of probe and gallery for testing, the temporal pooling method and other parameter settings, such as embedding size of RNN and margin $\alpha$ in Eqn. ~\eqref{con_loss}. In this paper, we regard the method in \cite{mclaughlinrecurrent} as our baseline, in which  the spatial-temporal features are also employed in a recurrent way but without two-stream structure or spatial fusion method, and all the optical flow maps serving as motion information are pre-extracted in off-line way. 
 \subsubsection{Effect of Different Motion Information }
 As described in Sec. \ref{motion_net}, our motion network  is able to end-to-end learn motion information from video sequence frames. Thus except for the Lucas-Kanade optical flow (LK-Flow) algorithm~\cite{lucas1981iterative} used in \cite{mclaughlinrecurrent}, we  also exploit the optical flow maps produced by EpicFlow\cite{revaud2015epicflow} algorithm to investigate the effect of different motion information. The experimental results are shown in Tab. \ref{tab:lk_epic}. The AMOC networks in ``AMOC + LK-Flow'' and ``AMOC + EpicFlow'' methods are the non-end-to-end versions, the motion networks of which are replaced by the corresponding optical flow maps pre-extracted. Note that here we only study the effect of different optical flows carrying motion context information and verify the effectiveness of end-to-end learning. Therefore, all the shown results of our AMOC are achieved by using the ``concatenation fusion'' method introduced in the Sec. \ref{sp_fuse} and fusing two-stream networks at ``Max-pool2'' layer as illustrated in Fig.~\ref{fig:fusion}. In Tab.~\ref{tab:lk_epic}, ``Baseline + LK-FLOW'' is the method from \cite{mclaughlinrecurrent}, and we observe that the performance is boosted when the LK-Flow is replaced by the EpicFlow to produce optical flow in either baseline methods or our AMOC method. As introduced in \cite{revaud2015epicflow}, EpicFlow assumes that contours often coincide with motion discontinuities and computes a dense correspondence field by performing a sparse-to-dense interpolation from an initial sparse set of matches, leveraging contour cues using an edge-aware distance.  So it is more robust to motion boundaries, occlusions and large displacements than the LK-Flow algorithm\cite{lucas1981iterative}, which is beneficial to the extraction of motion information between video frames. In addition, we can see that the performance of our ``end-to-end AMOC'' using different optical flow generation methods is both improved. 
 
 \begin{figure*}[htb]
 	\centering
 	\includegraphics[width=14cm]{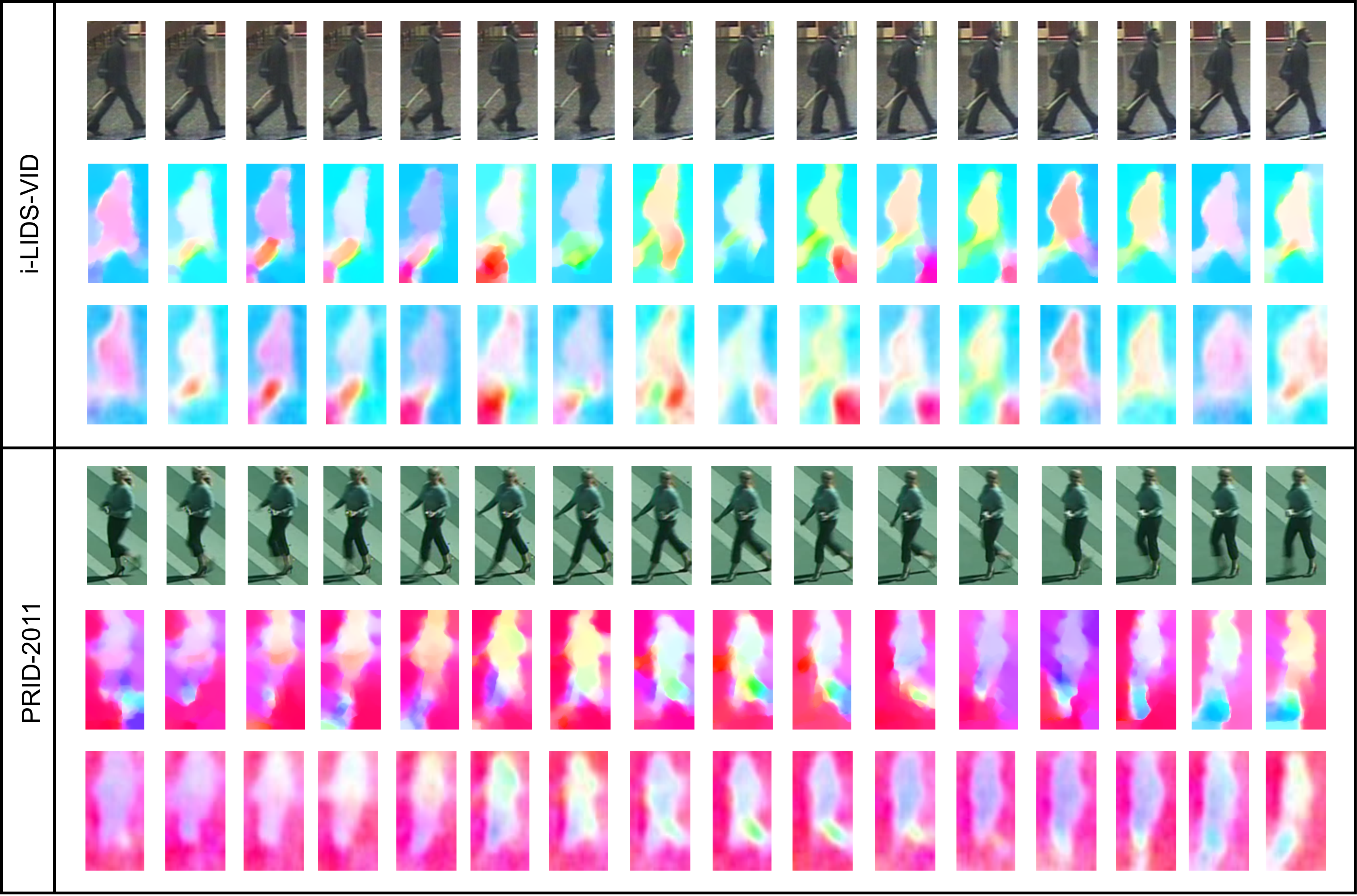}
 	\vspace{-3.5mm}	
 	\caption{The visualization of output optical flow of the proposed motion networks performed on iLIDIS-VID and PRID-2011 datasets. The first and forth rows are the consecutive raw video frames from iLIDS-VID and PRID-2011 datasets while the optical flow maps computed using EpicFlow\cite{revaud2015epicflow} for the two datasets are shown in the second and fifth rows. The third row and sixth row are the output flow maps of our motion networks. All the produced optical flow maps are encoded exploiting the flow colour coding\cite{baker2011database} method. Different colours represent different directions of motions, shades of which indicate the speeds of motions. }
 	
 	%\vspace{-5.5mm}	
 	\label{fig:flow_demo}
 \end{figure*}

 \begin{figure}[htb]
 	\begin{center}
 		\begin{tabular}{cc}
 			{\hspace{-3pt}}
 			\includegraphics[height=0.39\linewidth, width=0.46\linewidth]{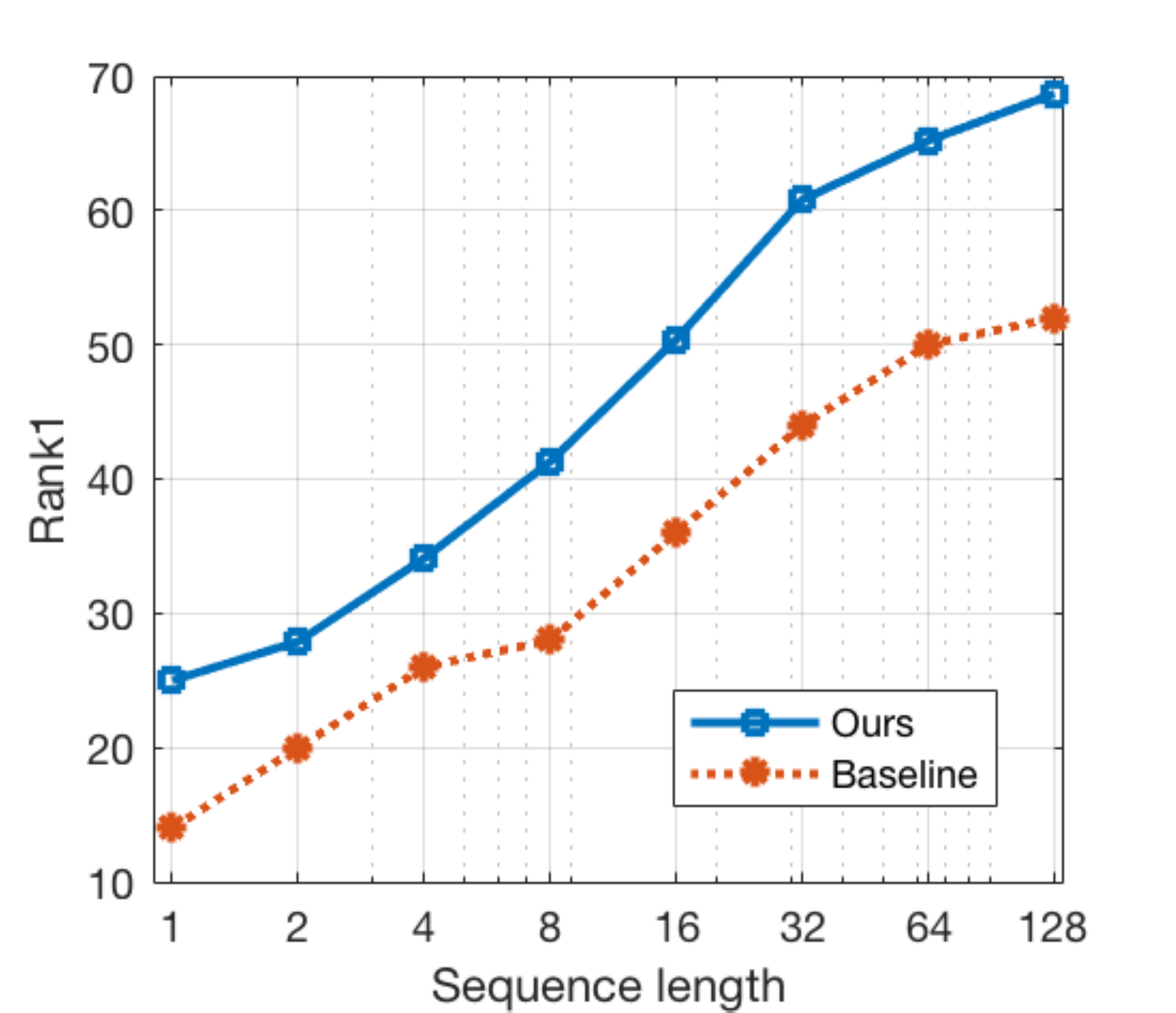}	&
 				\includegraphics[height=0.39\linewidth, width=0.46\linewidth]{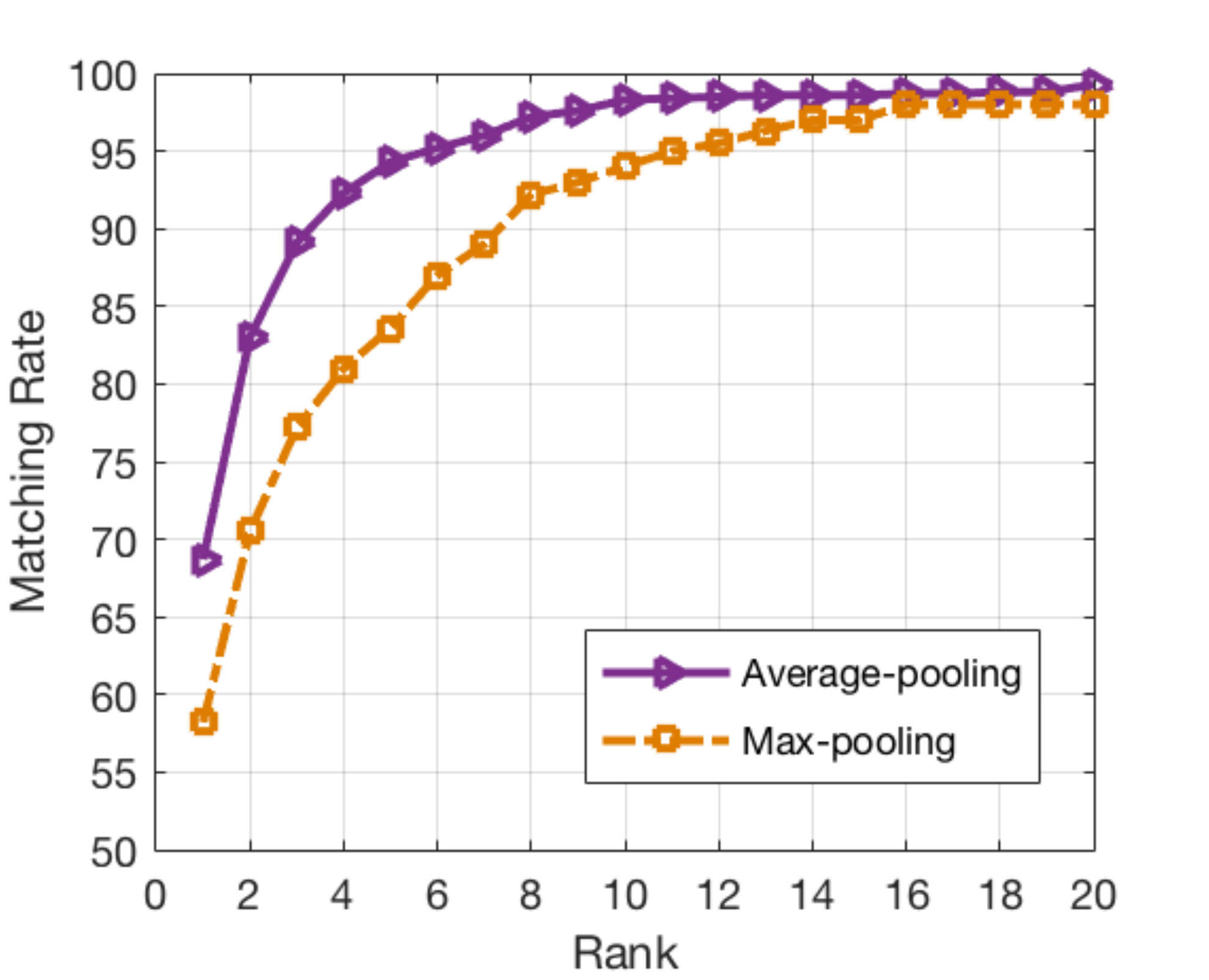}\\
 			{\footnotesize{(a) Effect of sequence length }} &
 		{\footnotesize{(b) Effect of pooling methods}} \\
 			\includegraphics[height=0.39\linewidth,width=0.46\linewidth]{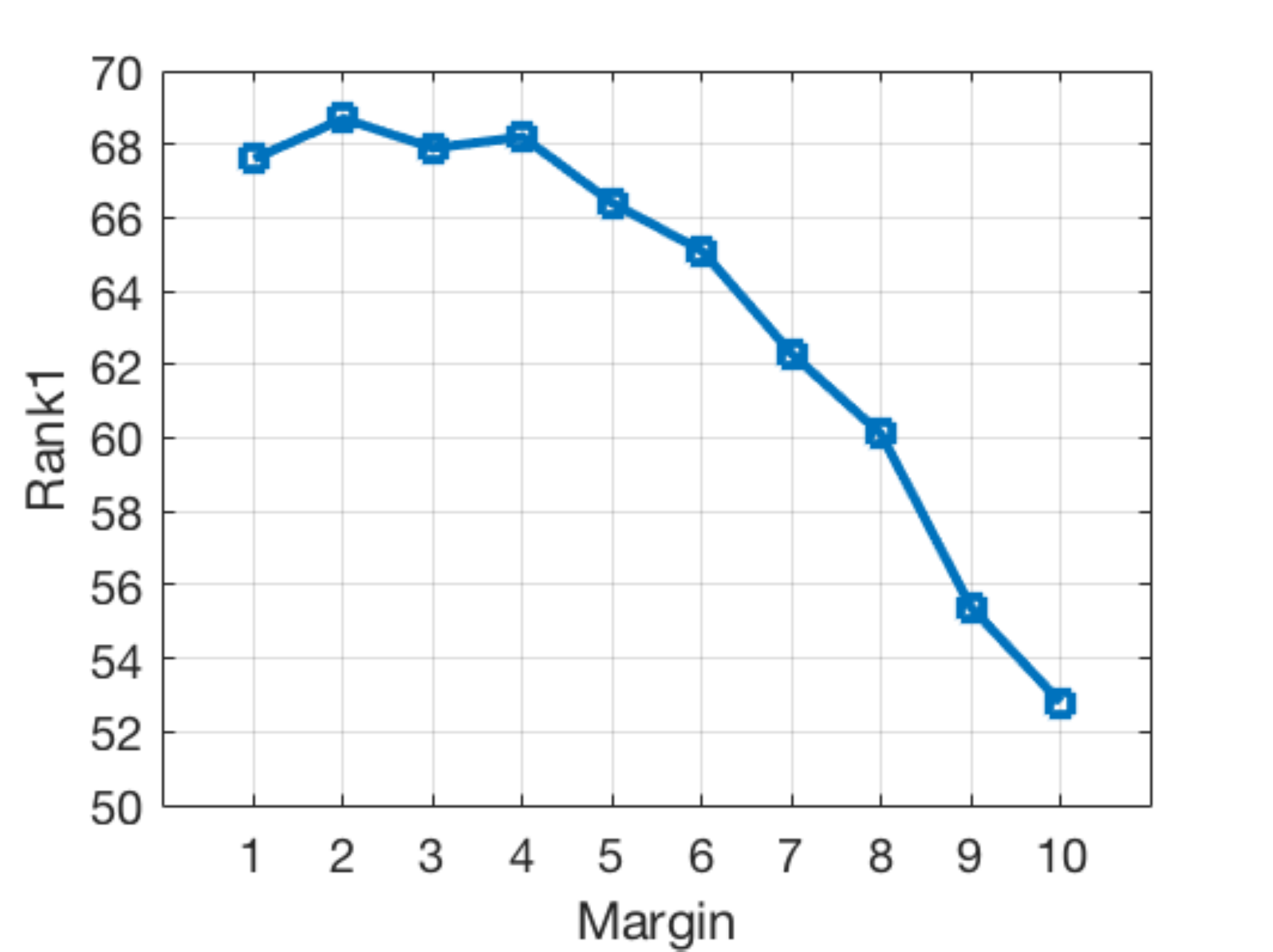}&
 			\includegraphics[height=0.39\linewidth,width=0.46\linewidth]{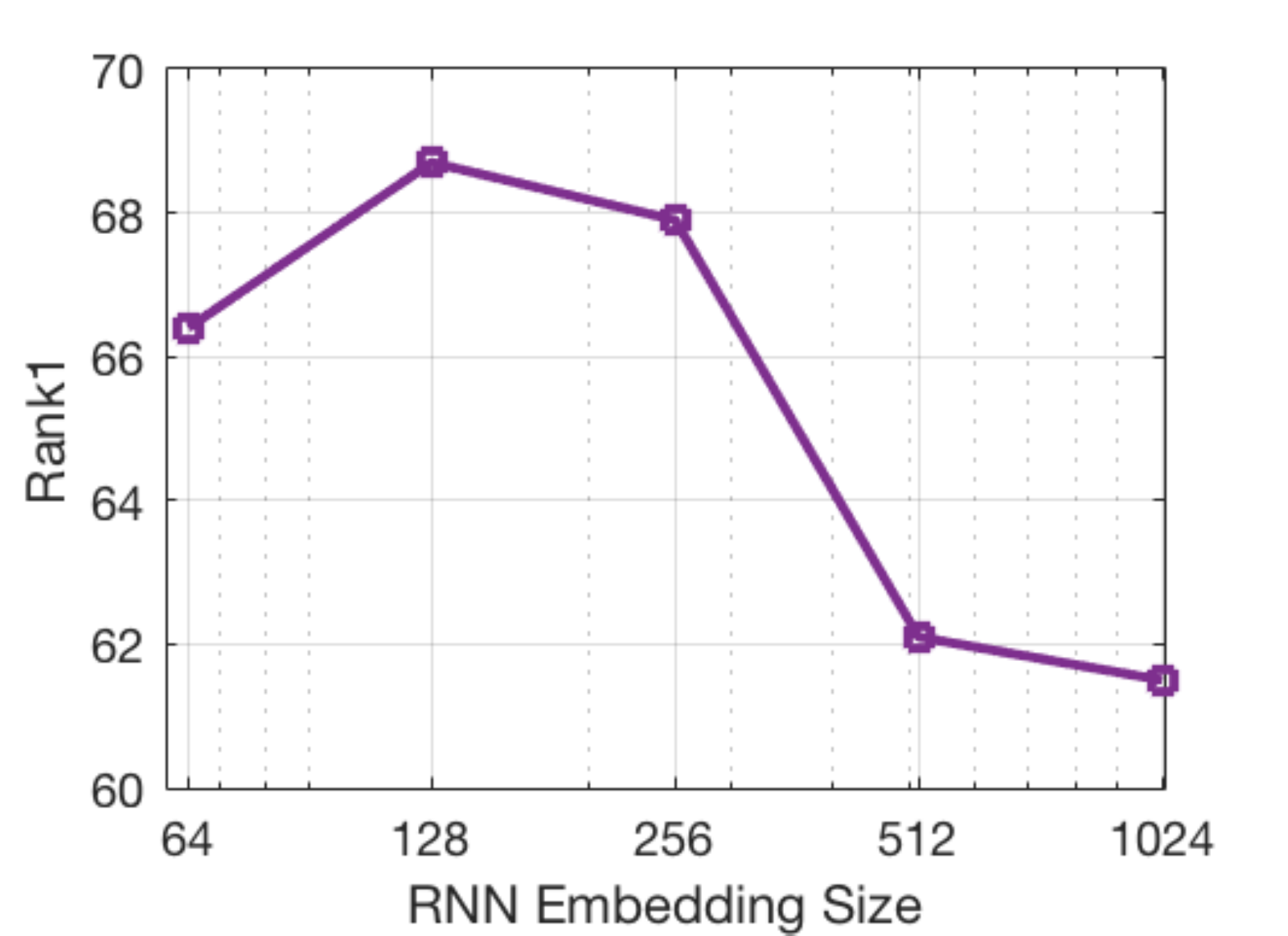}
 			\\
 		
 				{\footnotesize{(c) Effect of margin}}&
 				 {\footnotesize{(d) Effect of embedding size}}
 			
 			%{\vspace{0.5em}}
 		\end{tabular}
 	\end{center}
 	\caption{The performance of our proposed end-to-end AMOC using different parameter settings on iLIDS-VID dataset. (a) shows comparisons of our proposed end-to-end AMOC with baseline method\cite{mclaughlinrecurrent} using different query and gallery sequence lengths, ranging from 1 to 128, for testing. (b) shows the CMC curves of our method using two different temporal pooling methods (\textit{i.e}.,  average-pooling and max-pooling). (c) gives our method's performance variation depending on different margins in contrastive loss (Eqn.~\eqref{con_loss}). (d) illustrates the effect of different RNN embedding sizes upon the performance of our method.}
 	
 	\label{fig:ana_fig}
 	\vspace{-1em}
 \end{figure}
 
 Moreover, we also visualize the optical flow maps produced by our motion networks in Fig.~\ref{fig:flow_demo}.  The first and forth rows are the consecutive raw video frames from iLIDS-VID and PRID-2011 datasets while the optical flow maps computed using \cite{revaud2015epicflow} for the two datasets are shown in the second and fifth rows. The third row and sixth row are the output flow maps of our motion networks. All the produced optical flow maps are encoded with the flow color coding~\cite{baker2011database} method which is also used  in \cite{revaud2015epicflow}. Different colors represent different directions of motions, and shapes indicate the speeds of motions. The faster the motion is, the darker its color will be.  In this experiment, we only use the sequences of half persons of iLIDS-VID for training, and perform the motion networks on the left half persons' sequences of iLIDS-VID and the whole PRID-2011 dataset. In other words, for the PRID-2011 dataset, there is no need to re-train the motion networks on it. From the results shown in Fig.~\ref{fig:flow_demo}, we can see that our motion networks can well approximate the optical flow maps produced by EpicFlow~\cite{revaud2015epicflow} and successfully capture the motion details of persons, such as the speed and amplitude of legs moving. Especially for the PRID-2011 dataset, our motion network can achieve the good optical flow estimation without using training data from PRID-2011, which means our motion network has a good generalization ability. 
 
 To further illustrate the effectiveness of motion information embedded in the AMOC architecture, we remove the temporal stream (corresponding to the green and blue boxes in Fig. ~\ref{fig:framework}) of AMOC to perform the analytic experiments. Namely, our AMOC degenerates to the single-stream network which only uses appearance features of persons and there is no motion information included and accumulated. The results are reported as ``AMOC w/o Motion'' in Tab. \ref{tab:lk_epic}. Compared with other methods using motion information in Tab. \ref{tab:lk_epic}, the performance of ``AMOC w/o Motion'' is the worst for both datasets. This suggests that the motion information is beneficial for improving the person re-identification performance in video sequence.

\subsubsection{Effect of Spatial Fusion Method and Location}
In this part, we investigate how and where to fuse two-stream networks in our end-to-end AMOC network. For these experiments, we use the same spatial network architecture introduced in Sec.~\ref{spatial}. The fusion layer can be injected at any location, such as after ``Max-pool 2'', \textit{i.e.} its input is the output of ``Max-pool 2'' from the two streams. After the fusion layer a single processing stream is used. 

We compare different fusion strategies in Tab.~\ref{tab:fusion},  where we report the rank1, rank5, rank10 and rank20 recognition rate on both iLIDS-VID and PRID-2011 datasets. From the results, we see that ``Concatenation'' fusion method performs considerably higher than ``Sum'' and ``Max'' fusion methods. Compared to the fusion methods, our end-to-end AMOC network shows more sensitiveness to the location of spatial fusion. Specifically, for all three fusion methods, our method can achieve the best performance when the spatial fusion is performed after the ``Max-pool2'' layer. And fusion at FC (Fully-Connected) layers results in an evident drop in the performance. The reason for FC performing worse may be that at this layer spatial correspondences between spatial appearance and motion context information would be collapsed.

\begin{table*}[htbp]
	\linespread{1.3}\selectfont
	\centering
	\caption{Rank1, Rank5, Rank10 and Rank20 recognition rate (in \%) of various fusion methods on iLIDS-VID and PRID-2011 datasets. }
	\begin{tabular}{c|c|cccc|cccc}
		\hline
		\multirow{2}{*}{\textbf{Fusion Method}} &\textbf{Dataset} &\multicolumn{4}{c|}{\textbf{iLIDS-VID}}&\multicolumn{4}{c}{\textbf{PRID-2011}}\\
		\cline{2-10}
		&\textbf{Fusion Layer} & \textbf{Rank1} & \textbf{Rank5} & \textbf{Rank10} & \textbf{Rank20} & \textbf{Rank1} & \textbf{Rank5} & \textbf{Rank10} & \textbf{Rank20}\\
		\hline
		
		\multirow{7}{*}{Sum} &{Tanh1} & 60.8 & 90.1 & 91.7 & 96.5 &71.5 & 93.4&97.3 &98.1\\
		%\cline{2-10}
		&{Max-pool1} & 61.2 & 89.8 & 90.3 & 95.1 &72.6 & 92.8&96.2 &96.9\\
		&{Tanh2} & 65.5 & 91.8 & 95.6 & 96.5 &79.9 & 95.8&97.6 &97.8\\
		&{Max-pool2} & 67.8 & 93.4 & 96.5 & 98.3 & 80.6 & 96.6 & 97.8 & 99.2 \\
		&{Tanh3} & 63.9 & 91.8 & 96.7 & 97.7 &78.3 & 94.5&97.8 &98.2\\
		&{Max-pool3} & 65.0 & 92.8 & 97.9 & 98.4 &78.8 & 94.9&98.0 &99.1\\
		&{FC} & 60.7 & 88.1 & 93.2 & 94.3 &72.0 & 91.2&93.8 &94.9\\
		\hline
		
		\multirow{7}{*}{Max}&{Tanh1} & 60.6 & 90.0 & 91.9 & 97.2 &73.1 & 94.9 &97.2 &99.5\\
		%\cline{2-10}
		&{Max-pool1} & 61.0 & 91.2 & 93.1 & 97.8 &75.1 & 94.9&99.0 &99.5\\
		&{Tanh2} & 66.9 & 94.2 & 97.1 & 98.9 &81.2 & 97.3&98.5 &99.3\\
		&{Max-pool2} & 68.2 & 95.4 & 97.5 & 98.9 & 81.6 & 98.6 & 98.8 & 99.3 \\
		&{Tanh3} & 64.0 & 91.5& 96.4 & 97.3 &78.1 & 94.1&97.3 &98.2\\
		&{Max-pool3} & 64.3 & 93.1 & 98.2 & 98.3 &78.4 & 94.5&97.8 &99.0\\
		&{FC} & 61.2 & 89.1 & 93.6 & 94.9 &73.5 & 90.2&92.9 &94.7\\
		
		\hline
		\multirow{7}{*}{\textbf{Concatenation}} &{Tanh1} & 64.1 & 92.3 & 94.2 & 98.0 &74.8 & 95.5&98.4 &99.2\\
		%\cline{2-10}
		&{Max-pool1} & 64.8 & 92.0 & 94.1 & 98.3 &75.3 & 95.8&97.6 &99.2\\
		&{Tanh2} & \textbf{69.2} & 93.8 & 97.9 & 98.7 &83.0 & 97.9&99.0 &99.6\\
		&\textbf{Max-pool2} & 68.7 & \textbf{94.3} & \textbf{98.3} & \textbf{99.3} & \textbf{83.7} & \textbf{98.3} & \textbf{99.4} & \textbf{100} \\
		&{Tanh3} & 65.2 & 92.3 & 97.1 & 98.4 &80.0 & 96.3&99.4 &99.6\\
		&{Max-pool3} & 66.1 & 92.8 & 97.9 & 98.4 &80.0 & 96.9&99.8 &99.8\\
		&{FC} & 62.3 & 88.3 & 93.9 & 95.6 &73.2 & 91.3&94.4 &96.7\\
		\hline
	\end{tabular}%
	
	\label{tab:fusion}%
\end{table*}%

\subsubsection{Effect of  Sequence Length and Temporal Pooling Methods}\label{lenandtp}
To further study how re-identification accuracy varies depending on the lengths of the probe and gallery sequences during the test phase, we perform the experiments on the iLIDS-VID dataset. The testing lengths of the probe and gallery sequences are set to the same number and  simultaneously increased from 1 to 128, in steps corresponding with the powers-of-two. Training lengths are set to 16 as indicated in the Sec.~\ref{exp_setting} .

The results shown in Fig.~\ref{fig:ana_fig} (a) prove that increasing both the probe and gallery sequence lengths can bring the improvement of performance. This is reasonable because more appearance and motion information would be exploited if more samples for each person are available. The similar phenomenon is also verified by the baseline method\cite{mclaughlinrecurrent}. Comparing to it, our method can achieve higher rank1 recognition rates by significant margins under different sequence length settings.

As aforementioned in the Sec.~\ref{mo_acc},  the AMOC accumulates the RNN output spatial-temporal features into a single feature vector by apply either average-pooling or max-pooling over the temporal dimension. Now we compare the performance of the two pooling methods. Results are summarized in 
Fig.~\ref{fig:ana_fig} (b). We observe that average-pooling is superior than the max-pooling. One possible reason is that average-pooling consider all the time steps equally important in the decision, whilst  max-pooling only employ the feature value in the temporal step with the largest activation. Therefore, average-pooling over the temporal sequence of features can produce a more robust feature vector to compress and represent the person's appearance and motion information over a period of time.

%\begin{table}[htbp]
%	\centering
%	\caption{iLIDS-VID CMC re-identification accuracy as the lengths of the probe and gallery sequences are increased simultaneously}
%	\begin{tabular}{c|cccc|cccc}
%	\hline
%	\textbf{Dataset} &\multicolumn{4}{c|}{\textbf{iLIDS-VID}}&\multicolumn{4}{c}{\textbf{PRID-2011}}\\
%		\textbf{Length} & \textbf{Rank1} & \textbf{Rank5} & \textbf{Rank10} & \textbf{Rank20} \\& \textbf{Rank1} & \textbf{Rank5} & \textbf{Rank10} & \textbf{Rank20}\\
%		\hline
%		1 & 25.0    & 51.7  & 69.1 & 82.2 \\
%	2  & 27.9 & 53.2  & 71.5 & 82.3 \\
%	4  & 34.1  & 57.3 & 75.1 & 89.1 \\
%	8  & 41.3  & 68.1 & 78.3 & 89.1 \\
%	16  & 50.4  & 77.6 & 88.5 & 95.3 \\
%	32  & 60.8  & 89.2 & 93.4 & 97.4 \\
%	64  & 65.2  & 90.1 & 95.2 & 98.2 \\
%	128  & 68.7 & 94.3 & 98.3 &99.3 \\
%		\hline
%	\end{tabular}%
%	\label{tab:dif_len}%
%\end{table}%

\subsubsection{Effect of Other Parameter Settings}
In this subsection, we conduct experimental analysis on iLIDS-VID to investigate the effect of other parameter settings on our proposed AMOC: the margin $\alpha$ in Eqn.~\eqref{con_loss} and embedding size of RNN. 

In this paper, we adopt multi-task loss including contrastive loss and classification loss. The margin $\alpha$, in the contrastive loss function, can be set empirically, since the AMOC model can learn to adaptively scale the feature vectors proportional to $\alpha$. But the choice of the margin is still crucial to the performance. Therefore, we investigate a few different margins ranging from 1 to 10. The rank1 recognition rate is used for evaluation and the result curve is shown in Fig.~\ref{fig:ana_fig} (c). We observe that the rank1 recognition rate stays stable when we set margin smaller than 5. And our model can achieve the best rank1 performance (68.7\%) when the margin is set to 2. If we further increase the margin from 5 to 10, the performance would drop considerably. Therefore, 2 is the best choice of margin in our model. 

Additionally, the effect of different embedding sizes (\{64, 128, 256, 512, 1024\}) of RNN on the rank1 performance of our method  is also illustrated in the Fig.~\ref{fig:ana_fig} (d). We find our method can obtain the highest rank1 recognition rate under 128 embedding size setting. When the embedding size is reduced to 64, the rank1 drops slightly. This is probably because of information loss from the reduction of parameters. When we use 256 embedding size, the slight drop can also be observed. Moreover, the performance would be constantly undermined if the embedding size is further increased to 1024, mainly due to over-fitting. Besides, increasing the embedding size can also bring longer training time. So we choose the 128 embedding size of RNN as it gives the best trade-off between performance and computational cost.

 \subsection{Comparison with State-of-the-Art Methods}
 We further evaluate the performance of end-to-end AMOC, by comparing it with the state-of-the-art methods on iILIDS-VID, PRID-2011 and MARS datasets. The methods includes STA\cite{liu2015spatio}, DVR\cite{wang2016person}, TDL\cite{you2016top}, SI$^{2}$DL\cite{zhu2016video}, PaMM\cite{cho2016improving}, mvRMLLC+ST-Alignment\cite{chen2015person}, TAPR\cite{gao2016temporally}, SRID\cite{karanam2015sparse}, AFDA\cite{Li2015Multi}, DVDL~\cite{Karanam2015Person}, HOG3D\cite{Kl2008A}, KISSME\cite{K2012Large}, GEI\cite{Han2005Individual}, HistLBP\cite{xiong2014person} XQDA\cite{liao2015person}, LOMO\cite{liao2015person}, BoW\cite{zheng2015scalable}, gBiCov\cite{ma2014covariance}, IDE\cite{zheng2016person} and RFA-Net\cite{yan2016person}. Note, as discussed in Sec.~\ref{ana}, the motion network of our model has a good generalization ability. Therefore, for all the experiments of our end-to-end AMOC performed on all three datasets, we only use the  iLIDS-VID dataset to pre-train the motion network. 
 
 \begin{table*}
 	\linespread{1.5}\selectfont
 	\centering
 	\caption{Comparison of our end-to-end AMOC's performance on iLIDS-VID and PRID-2011 datasets to the state-of-the-arts.  }
 	\begin{tabular}{c|cccc|cccc}
 		\hline
 		\textbf{Dataset} &\multicolumn{4}{c|}{\textbf{iLIDS-VID}}&\multicolumn{4}{c}{\textbf{PRID-2011}}\\
 		\hline
 		\textbf{Methods} & \textbf{Rank1} & \textbf{Rank5} & \textbf{Rank10} & \textbf{Rank20} & \textbf{Rank1} & \textbf{Rank5} & \textbf{Rank10} & \textbf{Rank20}\\
 		\hline
 		Baseline \cite{mclaughlinrecurrent} & 58.0 & 84.0 & 91.0 & 96.0 & 70.0 & 90.0 & 95.0 & 97.0\\
 		STA\cite{liu2015spatio} & 44.3 & 71.7 & 83.7 & 91.7 &64.1 & 87.3&89.9 &92.0\\
 		DVR\cite{wang2016person} &39.5&61.1&71.7&81.8&40.0&71.7&84.5&92.2\\
 		TDL\cite{you2016top} &56.3&87.6&95.6&98.3&56.7&80.0&87.6&93.6\\
 		$\mathrm{SI^{2}DL}$\cite{zhu2016video}&48.7&81.1&89.2&97.3&76.7&95.6&96.7&98.9\\
 		PaMM\cite{cho2016improving} &30.3&56.3&70.3&82.7&56.5&85.7&96.3&97.0\\
 		mvRMLLC+ST-Alignment\cite{chen2015person}&\textbf{69.1}&89.9&96.4&98.5&66.8&91.3&96.2&98.8\\
 		TAPR\cite{gao2016temporally}&55.0&87.5&93.8&97.2&73.9&94.6&94.7&98.9\\
 		SRID\cite{karanam2015sparse}&24.9&44.5&55.6&66.2&35.1&59.4&69.8&79.7\\
 		AFDA\cite{Li2015Multi}   &37.5&62.7&73.0&81.8&43.0&72.7&84.6&91.9\\
 		DVDL\cite{Karanam2015Person} &25.9&48.2&57.3&68.9&40.6&69.7&77.8&85.6\\
 		RFA-Net\cite{yan2016person}&49.3&76.8&85.3&90.0&58.2&85.8&93.4&97.9\\
 		\hline
 		AMOC + EpicFlow & 65.5& 93.1& 97.2 & 98.7 &82.0 &97.3 &99.3 &99.4\\
 		\textbf{end-to-end AMOC + EpicFlow} & 68.7& \textbf{94.3} & \textbf{98.3} & \textbf{99.3} & \textbf{83.7} & \textbf{98.3} & \textbf{99.4} & \textbf{100} \\
 		\hline
 	\end{tabular}%
 	\label{tab:comp}%
 \end{table*}%
 
 \subsubsection{Results on iLIDS-VID and PRID-2011}\label{res_ilids}
 
 Comparing the CMC results shown in Tab.~\ref{tab:comp}, we can see that the non-end-to-end version of our AMOC can achieve higher performance than all the compared methods for both iLIDS-VID and PRID-2011 datasets. When the end-to-end AMOC is applied, the performance is further boosted, especially for the Rank1 protocol. The improvements are 3.2\% and 1.7\% for iLIDS-VID and PRID-2011 datasets respectively. Moreover, to our best knowledge, we are the first to introduce a two-stream network structure and end-to-end learning motion information from raw frame pairs  to solve the video-based person re-identification problem. Compared to those methods also using spatial-temporal features, such as STA~\cite{liu2015spatio}, RFA-Net\cite{yan2016person}  and TAPR\cite{gao2016temporally},  our AMOC can achieve performance improvement by a large margin for both datasets, due to the two-stream structure of our AMOC which separately deals with the spatial appearance and motion information from context and then performs feature integration through spatial fusion. From the results, we notice that the second best method, ``mvRMLLC+ST-Alignment\cite{chen2015person}'', can also achieve good performance on iLIDS-VID. However, this method needs the complex pre-processing of person video frames while our end-to-end AMOC can directly learn representations from raw video frames. Note, in Tab. \ref{tab:comp}, ``end-to-end AMOC + EpicFlow'' means our end-to-end AMOC uses the motion networks pre-trained by using EpicFlow optical flow maps as the supervision.

\subsubsection{Results on MARS}\label{res_mars}
MARS is a large and realistic video-based person re-id dataset since it was captured in the campus of  university with complex environment. Besides, it contains several natural detection/tracking errors as the person videos were collected by applying the automatic DPM detector and GMMCP tracker. Each person is captured by six cameras at most in MARS~\textemdash~compared with iLIDS-VID and PRID-2011, MARS has a much larger scale: 4 times and 30 times larger in the number of identities and total tracklets, respectively. Therefore, the relationships between person pairs are more complicated. 

In Tab. \ref{tab:mars}, we compare our results with the state-of-the-art methods on MARS. The compared methods include 8 descriptors (\textit{i.e.}, SDALF, HOG3D, HistLBP, gBiCov, GEI, LOMO, BoW and IDE) and 3 metric learning methods (\textit{i.e.}, DVR, KISSME and XQDA).    

Among all the recent video re-id methods, the best known rank1 accuracy is 65.3\% on MARS under single query setting, reported in \cite{zheng2016mars}. From Tab. \ref{tab:mars}, we can observe that our AMOC model can achieve the Rank1 recognition rate higher than the current best method (IDE\cite{zheng2016person}+XQDA) by 3\%. Although the rank5 performance of AMOC is slightly lower than ``IDE+XQDA'', it still achieves the mAP as high as 52.9\%. Note, the descriptor ``IDE'' in ``IDE+XQDA'' is obtained by fine-tuning the CaffeNet\cite{Krizhevsky2012ImageNet} on ImageNet- pretrained model. By contrast, our model is simply trained only on MARS from scratch. Besides, in all the compared methods, the extraction of descriptors and metric learning are separated as two individual processes. That is to say, the extraction of descriptors can not update parameters in the process of metric learning. In our proposed method, the motion feature and appearance feature are end-to-end learned jointly with the following feature accumulation part. As a result, the learned features contain enough discriminative information for the ultimate person re-id target.  To conclude, the experimental results show that our model performs comparable to other methods in more complex multi-camera person re-identification tasks, benefiting from the inherent end-to-end trainable appearance and motion information accumulation mechanism.

\begin{table}[htbp]
	\centering
	\caption{Comparison of our end-to-end AMOC method’s performance on MARS dataset to the state-of-the-arts.}
	\begin{tabular}{c|ccc|c}
		\hline
	
		\textbf{Method} & \textbf{Rank1} & \textbf{Rank5}& \textbf{Rank20}& \textbf{mAP}\\
		\hline
		SDALF\cite{farenzena2010person}+DVR\cite{wang2016person}  & 4.1 & 12.3 &25.1 &1.8\\
	    HOG3D\cite{Kl2008A}+KISSME\cite{K2012Large} &2.6&6.4&12.4&0.8\\
	    HistLBP\cite{xiong2014person}+XQDA\cite{liao2015person}	&18.6&33.0&45.9&8.0\\
		gBiCov\cite{ma2014covariance}+XQDA                                &9.2&19.8&33.5&3.7\\
	   GEI\cite{Han2005Individual}+KISSME  &1.2 &2.8 &7.4 &0.4\\
		LOMO\cite{liao2015person}+XQDA &30.7&46.6&60.9&16.4\\
		BoW\cite{zheng2015scalable}+KISSME &30.6&46.2&59.2&15.5\\
		IDE\cite{zheng2016person}  + XQDA  &65.3 &\textbf{82.0} &{89.0} &47.6\\
		IDE + KISSME &65.0	&{81.1}	&{88.9}	&45.6\\
		\hline
		\textbf{end-to-end AMOC+EpicFlow} & \textbf{68.3} & 81.4 &\textbf{90.6}&\textbf{52.9}\\
		\hline
	\end{tabular}%
	\label{tab:mars}%
\end{table}%

\subsection{Discussion}
We here discuss some potential limitations of our proposed model though its practical  effectiveness and superior performance have already been demonstrated in the above experiments.  

One potential issue with our model is the learned flow information may be redundant and not relevant to identifying the person of interest. From the visualization of optical flow maps estimated by the motion net of our model in Fig.~\ref{fig:flow_demo}, we observe that the foreground person and background present different colours (\textit{i.e.}, motion patterns)  in each frame. This is reasonable as the person and background move with different velocities in the video. As we explained in Sec.~\ref{pre_motion}, the motion net within our model is pre-trained using pre-extracted optical flow as supervision. If the supervision optical flow includes too much non-related motion information (\textit{i.e.}, background motion information),the person motion will be dominated by background motion within the computed optical flow. This is quite possible for the dataset including person video sequences automatically obtained by the detector and tracker, such as MARS, because inherent detection or tracking errors can bring more non-related motion information. Although deep networks that our method utilized are mostly powerful for learning robust feature representation, we believe the video person re-id performance would be further boosted if non-related motion information can be effectively suppressed like leveraging object segmentation models. We will study this issue in our future works.

\section{Conclusion}
In this work, we propose an end-to-end Accumulative Motion Context Network (AMOC) based method addressing video person re-identification problem through joint spatial appearance learning and motion context accumulating from raw video frames. We conducted extensive experiments on three public available video-based person re-identification datasets to validate our method. Experimental results demonstrated that our model outperforms other state-of-the-art methods in most cases, and verified that our accumulative motion context model is beneficial for the recognition accuracy in person matching.

\section*{Acknowledgment}
This work was supported in part by the National Natural Science Foundation of China under Grant 61371155, Grant 61174170, and Grant 61632007, and in part by the China Scholarship Council under Grant 201506690007. 

The work of Jiashi Feng was partially supported by National University of Singapore startup grant R-263-000-C08-133, Ministry of Education of Singapore AcRF Tier One grant R-263-000-C21-112 and NUS IDS grant R-263-000-C67-646.

% or


%\appendices

%\section{}

%\section*{Acknowledgment}


\ifCLASSOPTIONcaptionsoff
  \newpage
\fi

\end{document}